\documentclass{article}
\ifdefined\pdfminorversion\pdfminorversion=7\relax\fi
\usepackage[T1]{fontenc}
\usepackage{iclr2027_conference,times}

\usepackage{amsmath,amsfonts,bm}

\def\eqref#1{equation~\ref{#1}}

\def\1{\bm{1}}

\DeclareMathAlphabet{\mathsfit}{\encodingdefault}{\sfdefault}{m}{sl}
\SetMathAlphabet{\mathsfit}{bold}{\encodingdefault}{\sfdefault}{bx}{n}

\usepackage{graphicx,booktabs,tabularx,array,longtable}
\usepackage{xcolor}
\usepackage{tikz}
\usetikzlibrary{arrows.meta,positioning,calc}
\usepackage{fontawesome5} %
\usepackage{pifont}   %
\usepackage{colortbl} %
\usepackage{float}
\floatstyle{ruled}
\newfloat{algorithm}{tbp}{loa}
\floatname{algorithm}{Algorithm}
\floatstyle{plain}
\usepackage{algpseudocode}
\usepackage[breakable,skins]{tcolorbox} %
\newtcolorbox{promptbox}[1]{breakable,enhanced jigsaw,colback=black!3,colframe=black!35,
  colbacktitle=black!10,coltitle=black,boxrule=0.4pt,arc=1pt,left=5pt,right=5pt,top=3pt,bottom=3pt,
  fonttitle=\small\bfseries,title={#1},fontupper=\small,
  before upper={\setlength{\parindent}{0pt}\setlength{\parskip}{0.45em}\raggedright}}
\usepackage[pdfversion=1.7]{hyperref}
\usepackage{url}
\usepackage{xurl}

\definecolor{draftblue}{RGB}{30,72,130}

\newcommand{\ncell}{\textcolor{black!40}{--}} %
\newcolumntype{Y}{>{\raggedright\arraybackslash}X}
\newcommand{\methodname}{APPL}
\newcommand{\support}{\mathcal D_N}

\title{Agent Priors-guided Policy Learning}
\author{
  \textbf{Puming Jiang\thanks{Equal contribution.}\kern0.4em, Tianrun Hu\footnotemark[1]\kern0.4em, Haozhe Du, Yibo Li,}\\
  \textbf{Zhiwei Xue, Xinhu Li, Harold Soh}\\
  National University of Singapore\\
  \texttt{p.jiang@u.nus.edu}, \texttt{tianrun.hu@u.nus.edu}\\
  \texttt{harold@nus.edu.sg}
}
\iclrfinalcopy
\hypersetup{pdftitle={Agent Priors-guided Policy Learning},
  pdfauthor={Puming Jiang, Tianrun Hu, Haozhe Du, Yibo Li, Zhiwei Xue, Xinhu Li, Harold Soh},
  pdfsubject={Preprint},colorlinks=true,linkcolor=draftblue,citecolor=draftblue,urlcolor=draftblue}

\begin{document}
\maketitle
\lhead{Preprint}
\begin{abstract}
Robots that learn from a few demonstrations often require two forms of generalization. Compositional generalization recombines skills to solve new tasks, and skill generalization lets the learned policy behind each skill work in new situations. The two depend on each other, yet information is lost between composition and the skills it calls. Where a skill works is determined by the structure its policy is trained with, while composition sees the skill only through a separate description, such as a name, an instruction, or a symbolic operator, that omits this structure. Our key idea is to use each policy's structural prior as part of the interface between composition and the skill. A structural prior states what a behavior depends on, for example that a grasp depends only on the gripper's pose relative to the object. Built into training, it shapes where the policy generalizes; stated in language, it tells composition where the policy applies. 
We instantiate this idea in Agent Priors-guided Policy Learning (\methodname{}). A construction agent segments complete demonstrations into reusable skills, proposes several structural priors for each skill, and trains and verifies one policy per prior. A runtime agent then selects among these prior-specific policies and composes them toward  new task goals using their interfaces. Across MetaWorld and long-horizon ManiSkill tasks, \methodname{} improves out-of-distribution skill generalization and enables previously unseen skill compositions; ablating the interface information substantially reduces performance. These results support the use of training-time structural assumptions as a bridge between skill learning and skill composition.
\end{abstract}

\section{Introduction}
\label{sec:introduction}
Robots in homes and warehouses are asked to do far more than they were shown. Demonstrations are costly, so a robot typically learns from a few demonstrations of a few tasks, while deployment brings new goals, new orderings of familiar operations, and objects in new places. Handling this requires two kinds of generalization. \emph{Compositional generalization} solves new tasks by recombining learned skills, and \emph{skill generalization} lets each skill still execute when objects move or when it starts from the state another skill left behind. The two depend on each other. A task-level decision is only as good as its knowledge of what each skill can physically do, and a skill is only useful if it generalizes to the states that task-level decisions put it in. Information, however, is lost between the task level and the skill level. The task level may reason about goals and sequences without knowing where a learned skill actually works, and each skill is trained from a few demonstrations without the structure that the task level relies on.

Existing approaches connect the task level and the skill level in different ways. 
Vision-language-action models couple task semantics and low-level control in a single model~\citep{kim2024openvla,black2025pi05}, but the conditions under which learned behaviors generalize remain implicit, and robustness to layout and object shifts remains limited~\citep{fei2026liberoplus}.
Task and motion planning keeps the levels separate and connects them through symbolic preconditions and effects~\citep{garrett2021integrated}, which are specified by hand or learned from data~\citep{konidaris2018skills,silver2023predicate}. Within this line, SymSkill~\citep{shao2025symskill} co-invents predicates, operators, and skills from demonstrations so that one structure serves both levels, but that structure is drawn from a predefined family of relative-frame predicates and dynamical-system controllers. Agentic systems can call policies as tools and use language as the interface~\citep{ahn2022saycan,shi2025hirobot,zhang2026harness}. A skill name or instruction tells the agent what a tool is for, but not the physical conditions under which it works, so the agent cannot ground its decisions in the tool's actual ability, and transitions between skills remain a common source of failure~\citep{rui2026diagnosing}.
Effective generalization therefore strongly depends on the \textit{interface} between the task and skill levels.

\begin{figure}[t]
\centering
\input{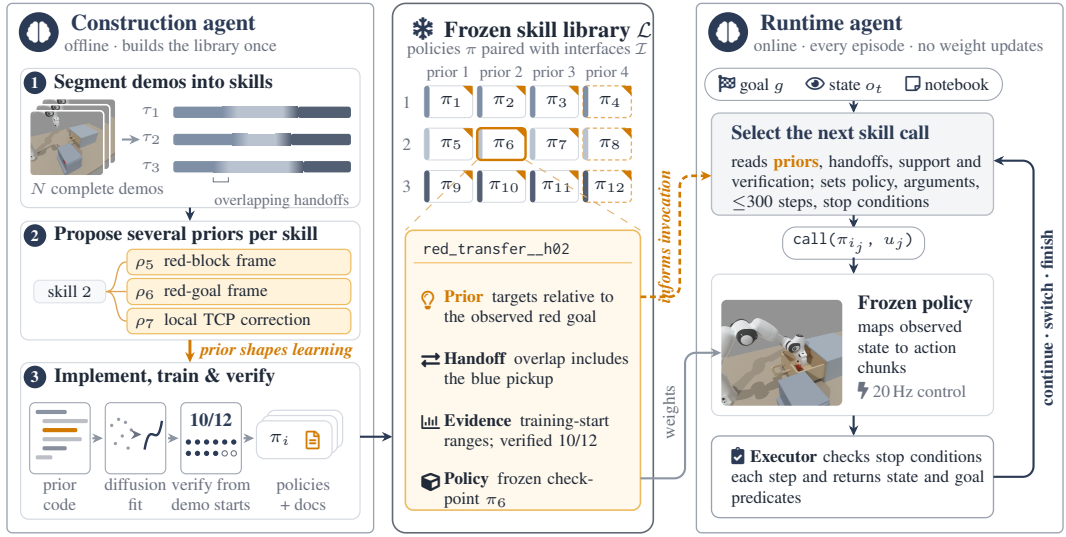}
\caption{\small\textbf{APPL builds prior-specific policies offline and composes them online.} The construction agent segments demonstrations, proposes several priors per skill, trains one documented policy per prior, and verifies each policy from demonstrated entry states. The runtime agent reads these documents to choose a policy, its arguments, a duration, and stop conditions. \textcolor[HTML]{D27A00}{Orange} marks the prior's two roles. %
} %
\label{fig:method}
\end{figure}

Our key idea is to \emph{use each skill's structural prior as part of this interface}. A structural prior specifies how a skill is designed to generalize, for example by expressing actions relative to the manipulated object rather than in absolute world coordinates. We use this prior in two ways. First, it is implemented in the skill through its representation or training objective, which shapes how the skill generalizes beyond its demonstrations. Second, the same prior is exposed to a task-level runtime agent, together with the skill's training support and handoff conditions. The agent can therefore reason not only about \emph{what} a skill does, but also about \emph{why and where} a particular implementation is expected to generalize. For example, if a grasp skill is trained in an object-relative frame, the agent can prefer it when the object appears at a new absolute location. The structural prior therefore serves both as an inductive bias for learning and as an explicit hypothesis about when the resulting skill should be applicable.

We instantiate this idea in \textbf{Agent Priors-guided Policy Learning} (\methodname{}, shown in Fig. \ref{fig:method}). A construction agent segments complete demonstrations into reusable skills and proposes several structural priors for each skill, rather than committing to one predefined abstraction or one policy implementation. It trains one Diffusion Policy~\citep{chi2023diffusion} for each prior and stores each policy together with a description of its prior, handoff conditions, observed training support, and verification evidence. The resulting library may therefore contain several implementations of the same skill whose different inductive biases make them suitable in different situations. At deployment, a runtime agent reads these descriptions to choose among the frozen policies, instantiate their arguments and stopping conditions, and compose them toward the task goal. In this way, information that shaped a policy during learning remains available when that policy is later selected and composed. \methodname{} thus turns the choice of policy structure from a fixed design decision into a per-skill hypothesis that can be proposed, implemented, and reused at runtime.

Our experiments test both roles of the structural prior. On six MetaWorld tasks, agent-designed priors substantially improve few-demonstration out-of-distribution skill generalization over vanilla Diffusion Policy and a fixed relational prior, reaching 89.6\% OOD success with two demonstrations versus 37.9\% for the fixed relational prior. On five long-horizon ManiSkill tasks with twelve demonstrations each, \methodname{} achieves 50.0\% success under shifted object configurations versus 10.0\% for full-task Diffusion Policy, 92.5\% on task-level variants with intermediate starts or requested early termination, and solves 8 of 16 previously unseen skill compositions. Importantly, when the same learned policy library is retained but its construction-time interface information is hidden from the runtime agent, success falls significantly. Together, these results support the central hypothesis of \methodname{} that the structural assumptions used to make a skill generalize can also provide useful information for deciding when and how that skill should be used.

\section{Related Work}
\label{sec:related_work}
\methodname{} targets compositional and skill generalization together by using each skill's structural prior as the interface between composition and the skill's policy.
We review work on composing learned skills, on structural priors for few-demonstration policies, and on language models that design parts of the learning pipeline.
Appendix~\ref{app:related} discusses further work and compares \methodname{} with the closest methods in detail.

\textbf{Composing learned skills.}
Task and motion planning composes skills through symbolic preconditions and effects~\citep{garrett2021integrated}, and a long line of work learns such abstractions from data as symbols grounded in skills, operators, or invented predicates~\citep{konidaris2018skills,silver2023neurosymbolic,silver2023predicate,liang2025visualpredicator}.
Skills with symbolic interfaces can also be learned through planner-guided reinforcement learning~\citep{cheng2023league} or from demonstrations~\citep{liu2025blade}, and skill boundaries can be discovered from unsegmented trajectories~\citep{zhu2022buds,wan2024lotus}.
Chaining fails when one skill ends in a state its successor does not support, and transition policies and skill-chaining methods address these handoffs~\citep{lee2019composing,lee2022tstar,mishra2023gsc}.
SymSkill is the closest precedent to \methodname{}~\citep{shao2025symskill}.
It jointly learns predicates, operators, and dynamical-system skills in relative frames from unsegmented demonstrations, and a symbolic planner composes and reorders these skills to reach new goals.
Its structure, however, is fixed in advance to relative frames and dynamical-system controllers.
MaestroMotif and SCALAR share language skill specifications between training and composition~\citep{klissarov2025maestromotif,zabounidis2026scalar}, but these specifications describe what a skill achieves rather than the inductive bias of its policy.
Language-model agents compose skills more flexibly~\citep{ahn2022saycan,huang2022inner,shi2025hirobot}, yet they know each skill only by its name or a feasibility estimate, and transitions or handoffs between skills can fail~\citep{rui2026diagnosing}.
\methodname{} instead lets an agent choose a different prior for each skill and keep alternative policies, and the prior that shapes each policy also tells the runtime agent where that policy applies.

\textbf{Structural priors for few-demonstration policies.}
Policies learned from few demonstrations, such as diffusion policies~\citep{chi2023diffusion}, tend to reproduce demonstrated behavior and can generalize poorly to out-of-distribution (OOD) states~\citep{he2026demystifying}.
Building task structure into the policy helps, for example through object-centric representations~\citep{zhu2023groot}, functional correspondence~\citep{tang2025functo}, affordances and contact locations~\citep{deng2026grow2,zhou2023hacman}, equivariance and task-relative frames~\citep{wang2024equivariant,rana2024oriented}, and visual cues~\citep{dai2025aimbot}.
Each such prior encodes an assumption that suits one family of operations, and a human chooses it for each task.
The prior is also used only in training and not in the system that later composes the policy.
Vision-language-action models obtain breadth from large-scale data rather than explicit structure~\citep{kim2024openvla,black2025pi05}, yet they remain sensitive to changes in layout and object pose~\citep{fei2026liberoplus}.

\textbf{Language models that design learning systems.}
Language models already design parts of robot learning pipelines, including reward functions~\citep{ma2024eureka}, simulation tasks and training data~\citep{wang2024robogen,wang2024gensim}, state abstractions~\citep{peng2024lga}, and keypoints~\citep{fang2025kalm}.
LGA and KALM are closely related to our work, since they let a model choose representation priors for imitation learning.
Each automates one kind of prior, produces one design per task, and uses that design only to train the policy. In contrast, 
\methodname{}'s construction agent proposes several kinds of priors for each skill, keeps the resulting policies as alternatives, and reuses each prior as the description over which the runtime agent composes skills.

\section{Problem Formulation}
\label{sec:formulation}

\textbf{Setting and objective.}
A robot acts with observations $o\in\mathcal O$ and actions $a\in\mathcal A$, and $\operatorname{Succ}(g,o)\in\{0,1\}$ indicates whether goal $g$ holds at $o$. A \emph{skill} $k$ is a reusable operation with a sub-goal $\beta_k$, such as opening a drawer, and a \emph{policy} $\pi_i$ implements skill $k(i)$. A system $\mathcal S=(\mathcal C,\mathcal L)$ consists of a library $\mathcal L=\{(\pi_i,\mathcal I_i)\}_{i=1}^{K}$, possibly with several policies per skill, and a composer $\mathcal C$ that solves a task by calling policies in sequence, seeing each only through its interface $\mathcal I_i$. Construction receives $N$ complete, unsegmented demonstrations $\support$ with goals in $\mathcal G_{\rm train}$ and task knowledge $\mathcal K$, and outputs a frozen system $\mathcal S$ that maximizes expected success,
\begin{equation}
J(\mathcal S)=\mathbb E_{g\sim P_{\rm test},\,o_0\sim\mu_{\rm test}(g)}\big[\operatorname{Succ}(g,o_T)\big].
\label{eq:objective}
\end{equation}

\textbf{Skill and compositional  generalization.}
Let the \emph{success region} $W(\pi_i)\subseteq\mathcal O$ be the set of start observations from which $\pi_i$ achieves $\beta_{k(i)}$ with probability at least $1-\epsilon$. \emph{Skill generalization} requires $W(\pi_i)$ to contain start states absent from the demonstrations, such as shifted objects or states left by a different preceding skill. Changes in object position with the task and skill sequence fixed test this ability; we call this setting \emph{motion-level OOD}.

\emph{Compositional generalization} requires solving tasks $(g,o_0)$ through skill sequences or handoffs absent from the demonstrations. Intermediate skills may be skipped, skills reordered, or a task may begin with a sub-goal already satisfied but with entry conditions for the next skill that differ from the demonstrated handoff. For example, putting the blue block into the drawer while leaving the red block inside skips the demonstrated removal of the red block and changes the placement skill's entry conditions. We separately evaluate \emph{task-level OOD}, which changes where a demonstrated task begins or is requested to end: resuming near a demonstrated stage or terminating at a requested intermediate sub-goal after a prefix of the demonstrated sequence. These cases test adapting execution to the current progress and requested endpoint while preserving the demonstrated skill order. Object displacement alone does not constitute a new composition.

Reliable composition requires the calls' sub-goals to jointly achieve $g$ and each call $j$ to start inside the success region of its policy, $o_{t_j}\in W(\pi_{i_j})$. The two kinds of generalization are thus coupled: a new sequence or handoff can also create start states that skill generalization must cover.

\textbf{The interface and desired properties.}
The composer cannot observe $W(\pi_i)$ in general. Instead, the interface $\mathcal I_i$ states an applicability region $A_i\subseteq\mathcal O$, and the composer can check only $o_{t_j}\in A_{i_j}$. Information is lost whenever $A_i$ and $W(\pi_i)$ differ. If $A_i\not\subseteq W(\pi_i)$, composition calls a policy where it fails. If $A_i$ is much smaller than $W(\pi_i)$, composition gives up usable calls. 
A useful interface is therefore \emph{faithful}, $A_i\subseteq W(\pi_i)$, and \emph{informative}, $A_i\approx W(\pi_i)$.
The problem is to construct, from $\support$ and $\mathcal K$ alone, policies whose success regions cover the states that new compositions create, together with interfaces that describe those regions faithfully and informatively.

\section{Agent-Guided Prior Design and Skill Composition}
\label{sec:method}

We present \methodname{}, which connects skill learning and skill
composition through structural priors. The central idea is to use the
same structural assumption in two roles: to shape how a skill policy
generalizes during training, and to describe when that policy is
expected to be applicable at runtime.

Figure~\ref{fig:method} summarizes our system. Construction happens
offline in stages. First, a construction agent segments complete
demonstrations into reusable skills. Next, for each skill it proposes
several structural priors, implements and trains one policy per prior. It then verifies the trained policies on demonstrated entry states. Finally, it writes the interface for each candidate and freezes the resulting library. At runtime, a separate agent reads these
interfaces to choose among the frozen policies and compose them toward
a new task goal. %

\subsection{Structural priors in the interface}
\label{sec:method_prior}

Different policies can fit the same demonstrations while having different success regions $W(\pi_i)$. We use a \emph{structural prior} $\rho_i$ to encode
additional assumptions about the relations the skill should depend on
and how its behavior should respond to changes in the scene. For
example, an object-relative prior expresses the assumption that a
manipulation behavior can be reused at different absolute object
locations when the relevant object-relative geometry is preserved.

\methodname{} uses each prior in two complementary ways. Its
\emph{training realization} shapes the learned policy. A prior may be
implemented through the policy representation, action parameterization,
architecture, or an auxiliary training objective. We write the
corresponding policy class as $\Theta_{\rho_i}$ and train
\begin{equation}
\theta_i
=
\arg\min_{\theta\in\Theta_{\rho_i}}
\;
\mathbb E_{\mathcal D_{k(i)}}\big[\ell_{\rm BC}(\theta)\big]
+
\lambda_i\,\ell^{\rm aux}_{\rho_i}(\theta).
\label{eq:train}
\end{equation}
The prior thereby shapes how the policy behaves beyond its
demonstrations and, consequently, its success region $W(\pi_i)$. We
refer to this role as \emph{coverage}.

The same prior also has a \emph{description realization}, which forms
part of the runtime interface $\mathcal I_i$. The prior and the observed
training support together provide a hypothesis about where the policy
should apply. For a representation-based prior, let
$\phi_{\rho_i}(o)$ denote the relevant transformed observation and let
$\mathcal R_i$ denote the region covered by the transformed skill
demonstrations. We define the corresponding hypothesized applicability
region as
\begin{equation}
A_i
=
\{o:\phi_{\rho_i}(o)\in\mathcal R_i\}.
\label{eq:support}
\end{equation}
For example, under an object-relative prior, a state with an object at
a new absolute location may still lie in $A_i$ when its relative
configuration lies within the demonstrated range. The interface
communicates this hypothesis through the prior description, measured
training support, and handoff conditions. The runtime agent uses this
information to decide which policy to invoke, which we refer to as
\emph{selection}.

This construction directly connects to the interface criteria of
Section~\ref{sec:formulation}. Using the same prior for training and
for the runtime description is intended to improve interface
faithfulness, $A_i\subseteq W(\pi_i)$, because the stated
applicability is based on the assumptions that shaped the policy.
This relationship is not guaranteed since it depends on both the validity
of the prior and how well the trained policy realizes it.
Informativeness additionally requires $A_i$ to capture as much of
$W(\pi_i)$ as possible rather than being unnecessarily conservative.

The prior therefore couples the two parts of the problem: its training
realization determines \emph{coverage}, while its description
realization supports \emph{selection}. Since the appropriate structure
can differ across skills and situations, construction must choose both
a segmentation $\sigma$ and priors for the resulting skill policies:
\begin{equation}
\max_{\sigma,\,\{\rho_i\}}
\;
J\big(\mathcal S(\sigma,\{\rho_i\},\support)\big).
\label{eq:design}
\end{equation}
This optimization ranges over an open-ended design space of policy
implementations and interfaces, and the deployment objective $J$ is
unavailable during construction. \methodname{} therefore uses a
language-model agent to propose, implement, and evaluate candidate
designs from the demonstrations and task knowledge.

\subsection{Segmenting demonstrations into skills}
\label{sec:method_segment}

The construction agent receives the $N$ complete demonstrations, the
task goals, and the observation and control conventions in
$\mathcal K$. It proposes skill boundaries on every trajectory and
groups segments that perform the same reusable operation into skill
datasets $\mathcal D_k$.
Adjacent skill segments are intentionally overlapped around their transition or handoff. In particular,
a skill's training data include part of the end of its predecessor and
the beginning of its successor. This overlap broadens the training
support around handoff states, allowing a successor policy to take
over from states that its predecessor actually reaches, including
states before nominal completion. The overlap reuses transitions from
the original demonstrations and introduces no additional demonstration
data.

\subsection{Proposing, implementing, and documenting priors}
\label{sec:method_implement}

\textbf{Proposing alternative priors.}
For each skill, the construction agent proposes several priors from
different families rather than committing to a single design. It is generally not possible to know
which structural assumption will generalize best to every state in
which the skill may later be invoked. As such, retaining several prior-specific
policies gives the runtime agent alternative implementations of the
same skill.

\textbf{Implementing the training realization.}
For each proposed prior $\rho_i$, the agent writes an implementation
against a fixed policy interface. The implementation constructs inputs
from the causal observation history, encodes demonstrated actions in
the coordinates specified by the prior, decodes predicted actions back
into native commands, and may introduce the auxiliary loss
$\ell^{\rm aux}_{\rho_i}$ from Equation~\ref{eq:train}. Each prior
 produces a separate conditional diffusion
policy~\citep{chi2023diffusion}, trained using a common recipe.
For example, consider a prior that represents actions in the observed
object frame $(R_t,p_t)$ at the start of an action chunk. End-effector
targets are transformed according to
\begin{equation}
x^{\rm loc}=R_t^\top(x^{\rm world}-p_t),\quad
\Delta x^{\rm loc}=R_t^\top\Delta x^{\rm world},\quad
\Delta x^{\rm world}=R_t\Delta x^{\rm loc}.
\label{eq:frame}
\end{equation}
This parameterization encourages the learned behavior to depend on
object-relative rather than absolute motion. In our experiments, a
shared inverse-kinematics module converts decoded end-effector targets
into native joint commands. %

\textbf{Verifying policy realizations before freezing.} 
For each candidate skill policy, the construction agent gathers limited
execution evidence. In our experiments, this  verification step uses only demonstrated skill-entry
states. Each policy is initialized from the entry state of its
corresponding segment in every demonstration, executed for 1.5 times
the segment length, and evaluated against the demonstrated exit
condition at the final state. The construction agent receives these
results and writes a verification report that separates observed
outcomes from its inferences about the policy. 
The agent then proposes one additional prior for each skill using this feedback. The corresponding policy is trained and verified in the
same way, after which the agent writes a usage note for every policy.

\textbf{Constructing the runtime interface.}
The construction agent then records an interface for every trained
skill policy,
\begin{equation}
\mathcal I_i=(d_i,\,h_i,\,s_i,\,v_i).
\label{eq:interface}
\end{equation}
The four fields provide complementary information about the skill.
The prior description $d_i$ states the structural assumption behind
$\rho_i$, including which observations and relations the policy was
designed to depend on, which variations it is expected to tolerate,
and known limitations. The handoff description $h_i$ records entry and
exit conditions and the predecessor and successor states represented
by the overlapping segments. The support description $s_i$ summarizes
the demonstrated training support, such as ranges of skill-entry
states. Finally, $v_i$ stores evidence gathered during verification. 
Together, these fields describe the policy's hypothesized applicability
region $A_i$ and the evidence available for supporting that hypothesis.
Each interface also declares the typed
arguments required by the policy, such as the object to manipulate and
its destination. %

\subsection{Composing skills at runtime} 
In our implementation, the policy library $\mathcal L$, its interfaces,
and the runtime agent's model, prompt, and tools are frozen after construction. 
At each decision point during execution, the runtime agent receives the task goal $g$,
the current state and goal predicates, recent interaction history, its
episode notebook, and the available policy interfaces $\mathcal I_i$.
It can inspect the full interface of any policy before invoking it.
Using this information, the agent chooses a policy and its call
arguments $u_j$. A call specifies the policy's typed arguments, an
execution duration, and stop conditions
over observed quantities, either absolute or relative to the start of
the call. For example, the agent may require both a gripper width below
3\,cm and the grasped object to be lifted by more than 4\,cm.

The runtime agent $\mathcal R_\alpha$ thereby instantiates the composer
$\mathcal C$ introduced in Section~\ref{sec:formulation}:
\begin{equation}
(i_j,u_j,m_{j+1})
\sim
\mathcal R_\alpha(
\,\cdot\mid h_{t_j},g,m_j,\mathcal L),
\qquad
\mathbf a_{t:t+H}
\sim
\pi_{i_j}(\,\cdot\mid h_t,u_j),
\quad
t_j\leq t<t_{j+1},
\label{eq:task_interface}
\end{equation}
where $h_t$ is the interaction history, $m_j$ is the episode notebook,
and $\mathbf a_{t:t+H}$ is an action chunk of $H$ steps. The executor runs the selected policy with fresh observations and
checks the specified stop conditions after every step. When a condition
fires, control returns to the runtime agent together with the updated
state and goal predicates. In our implementation, the executor also
interrupts a call immediately when the overall task goal is satisfied.
The agent can then continue the current skill, switch to another policy
for the same skill, invoke a different skill, or terminate the task. It
cannot write control code or modify the frozen library, and its memory
is reset between episodes.

Runtime composition is where the two roles of the structural prior
come together. The training realization determines the states from
which each policy can actually succeed, while the interface exposes
the construction-time hypothesis about that competence. The runtime
agent uses the latter to select and chain policies so that the library's
learned coverage can be exploited when solving new tasks.

\section{Experiments}
\label{sec:experiments}
This section reports on two experiments designed to evaluate APPL's ability to generalize. Specifically,  
Exp.\ 1 tests coverage in isolation: whether agent-designed priors make individual skills generalize. Exp.\ 2 tests coverage and selection together: whether priors, used both in training and as descriptions read by a runtime agent, let skills compose into long-horizon tasks when objects are shifted and when the requested task differs from every demonstrated one.

\subsection{Agent-designed priors for skill generalization}
\label{sec:motion_results}
Exp.\ 1 evaluates the training realization of priors and asks: given a skill and a few
demonstrations, can an agent design and implement priors that make the skill's
policy generalize? Skills and subgoals are supplied, so no composition is involved.

\textbf{Setting and comparisons.}
We use six adapted MetaWorld tasks~\citep{yu2020metaworld}: pick-place-wall,
assembly, drawer-open, door-open, peg-insert-side, and stick-push. Each has 20
successful demonstrations. 
Test
states vary two spatial factors per task: IID states stay within the demonstrated
combinations, C recombines the factors, and E places both outside their
demonstrated intervals; OOD weights C and E equally. All policies receive the
same state channels, and the agent may also inspect demonstration images while
designing. B0 is a vanilla diffusion policy~\citep{chi2023diffusion}, and B1 adds
a learned encoding of three predefined role-relative vectors; B1 is one fixed
relational prior, not an exhaustive set of expert priors. Agent candidates share
the diffusion backbone, the 20,000-update budget, the controller, and the
evaluation, but may change representations, encoders, action coordinates, and
auxiliary losses.

\textbf{Candidate design and selection.}
In each of the 24 task--$N$ conditions, the agent submits three candidates
(A1--A3) before any performance feedback, then designs a fourth (A4) from their
development results on fresh IID, C, and E states and trains it from scratch.
We report $q_1=\mathrm{A1}$, the first proposal, and $q_3$ and $q_4$, the
candidates with the best development OOD success among A1--A3 and A1--A4, with
fixed tie-breaks; the three share 144 trained models. All designs and selections
are frozen before testing, and neither IID nor test scores are used to select
candidates (training details in Appendix~\ref{app:exp1-training}).

\begin{table}[t]
\centering
\small
\caption{\small Exp.~1 hidden-test success (\%), averaged equally over six tasks, with $N$ demonstrations per task. IID uses fresh states within the demonstrated factor ranges; OOD averages recombined (C) and extrapolated (E) states equally. Mean averages OOD over the four values of $N$. Bold marks the best entry in each column. $q_1$ is the first proposal (A1); $q_3$ and $q_4$ are selected by development OOD success from A1--A3 and from A1--A4.}
\label{tab:exp1-main-ood}
\setlength{\tabcolsep}{3.4pt}
\renewcommand{\arraystretch}{1.12}
\begin{tabular}{l*{9}{c}}
\toprule
 & \multicolumn{2}{c}{$N=2$} & \multicolumn{2}{c}{$N=5$} & \multicolumn{2}{c}{$N=10$} & \multicolumn{2}{c}{$N=20$} & Mean \\
\cmidrule(lr){2-3}\cmidrule(lr){4-5}\cmidrule(lr){6-7}\cmidrule(lr){8-9}\cmidrule(l){10-10}
Method & IID & OOD & IID & OOD & IID & OOD & IID & OOD & OOD \\
\midrule
Diffusion Policy (B0) & 73.33 & 28.96 & 82.50 & 37.29 & 97.50 & 38.54 & 100.00 & 46.67 & 37.86 \\
Relational prior (B1) & 85.83 & 37.92 & 93.33 & 42.71 & \textbf{100.00} & 52.29 & 100.00 & 56.25 & 47.29 \\
\midrule
\methodname{}, first proposal ($q_1$) & 84.17 & 56.04 & 95.83 & 63.54 & \textbf{100.00} & 73.33 & 100.00 & 78.75 & 67.92 \\
\methodname{}, best of three ($q_3$) & 90.83 & 80.00 & \textbf{100.00} & 81.46 & \textbf{100.00} & 90.42 & 100.00 & 88.96 & 85.21 \\
\midrule
\rowcolor{black!6}\textbf{\methodname{} ($q_4$)} & \textbf{96.67} & \textbf{89.58} & \textbf{100.00} & \textbf{93.54} & \textbf{100.00} & \textbf{93.33} & 100.00 & \textbf{93.13} & \textbf{92.40} \\
\bottomrule
\end{tabular}
\end{table}

\textbf{Results.} Overall, the experiment results support the notion that agent provided priors improve skill generalization. 
Agent-designed priors raise six-task macro OOD success at every $N$
(Table~\ref{tab:exp1-main-ood}). At $N=2$, $q_4$ reaches 89.58\%, versus
28.96\% for B0 and 37.92\% for B1; its margins over B1 are 51.67, 50.83, 41.04,
and 36.88 percentage points for $N=2$, 5, 10, and 20. IID success is similar
across methods at larger $N$, so the gains come from generalization rather than
fitting. Intervals and per-task results are in
Appendices~\ref{app:exp1-statistics} and~\ref{app:exp1-full-results}.

\textbf{What the agent designs and what feedback adds.}
The first proposal alone beats B1 by 18.1--22.5 percentage points. Proposing
three priors and selecting by development OOD success adds 10.2--24.0 points,
and the feedback round adds 2.9--12.1 more: A4 is selected in 15 of 24
conditions, and test success rises in 10, ties in 13, and drops in one. Two
revisions show how this happens. At drawer/$N=2$, A1 expresses the hand--handle relation and
the actions in the cabinet frame (75.00\%). Because demonstrated planar actions
are close to $4(p^{\rm handle}-p^{\rm hand})$, A4 subtracts this term in the
action encoding, so the network predicts only a residual that extrapolates to
new positions, reaching 100.00\%. At assembly/$N=5$, the demonstrations
correlate nut and goal positions, and development feedback showed crossed
layouts failing. A4 attenuates goal features and the goal-aligned frame until
the ring is lifted, removing that shortcut, and rises from 41.25\% to 93.75\%.
The feedback covers recombined and extrapolated development states, which lets
revisions target the failing factor combinations. A revision can also hurt. For example, 
door/$N=20$ falls from 100.00\% under $q_3$ to 96.25\% under $q_4$. Because each
step from $q_1$ to $q_4$ adds candidates and development selection, $q_4-q_3$
does not isolate the effect of feedback (Appendix~\ref{app:exp1-mechanisms}).

\subsection{Long-horizon composition under shifted objects and unseen task variants}
\label{sec:composition_results}
\begin{figure}[t]
\centering
\setlength{\tabcolsep}{0.8pt}
\begin{tabular}{@{}ccccc@{}}
\footnotesize Drawer exchange & \footnotesize Buffer exchange & \footnotesize Retrieve and store & \footnotesize Covered peg assembly & \footnotesize Pour and return \\
\includegraphics[width=0.195\linewidth]{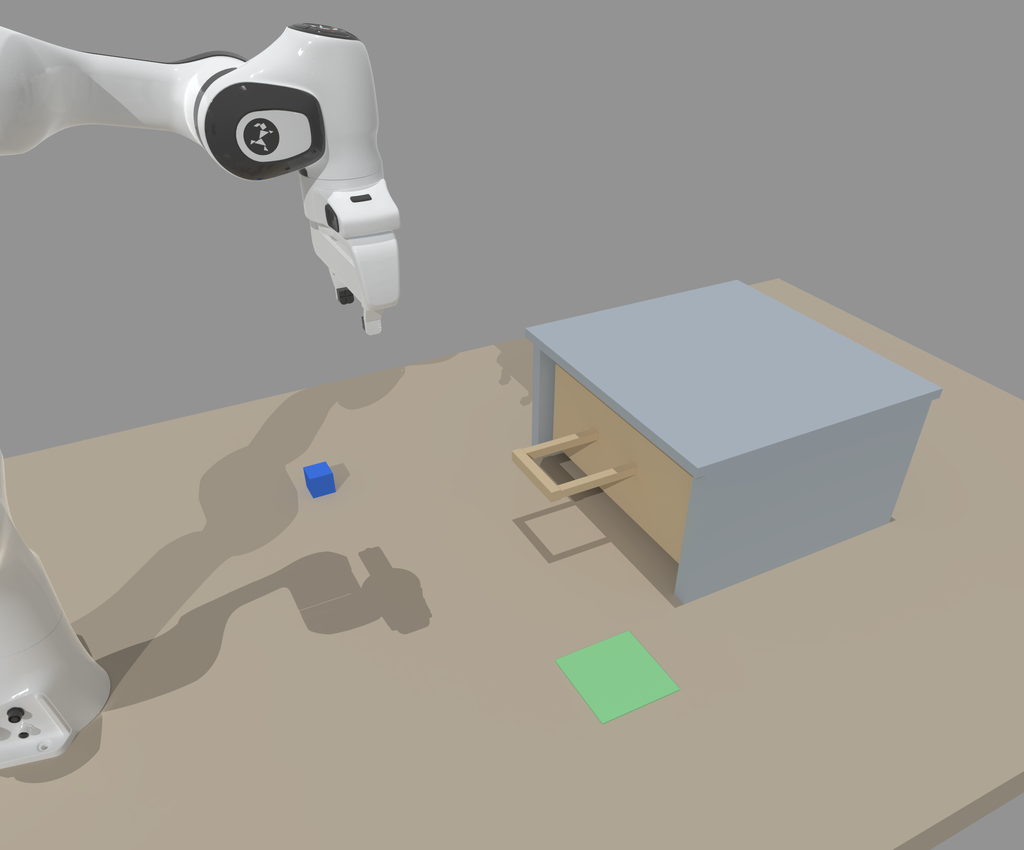} &
\includegraphics[width=0.195\linewidth]{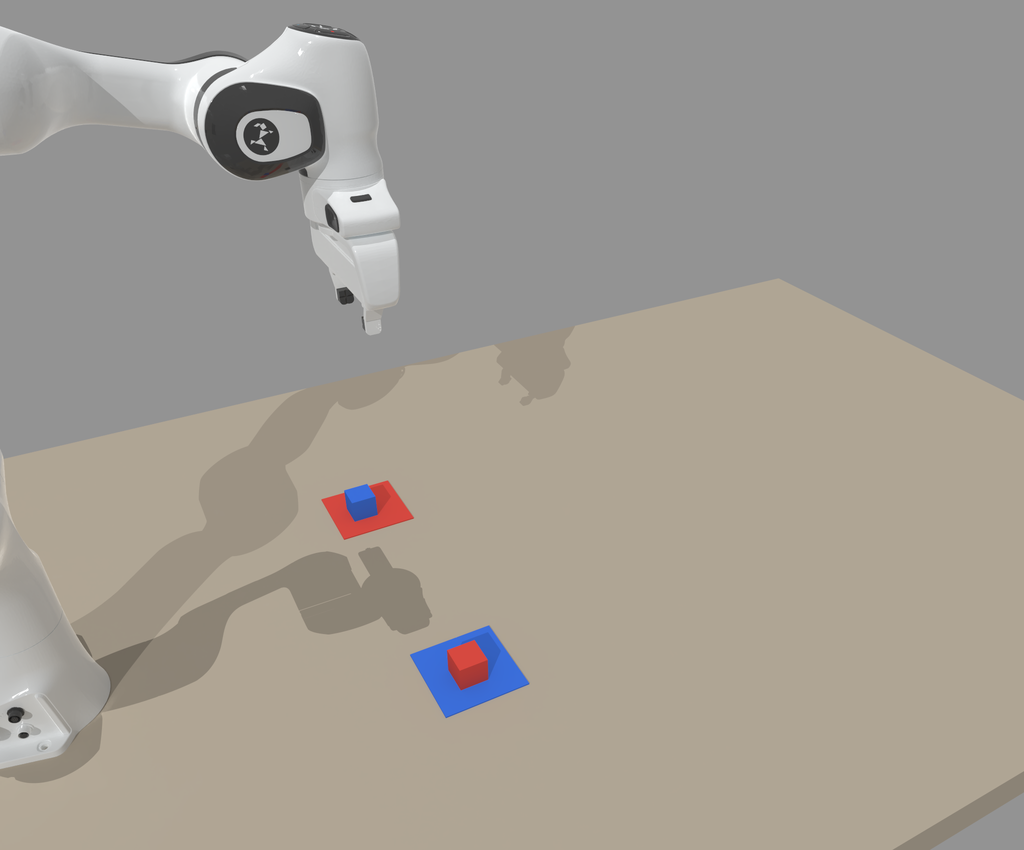} &
\includegraphics[width=0.195\linewidth]{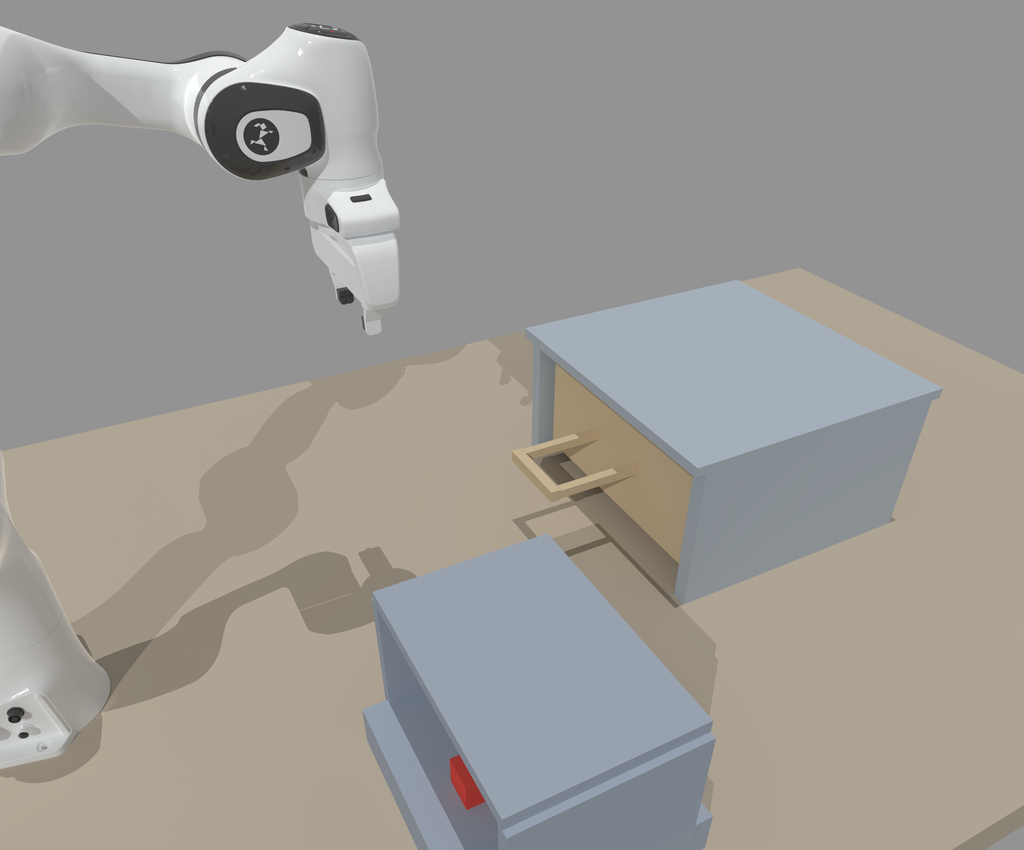} &
\includegraphics[width=0.195\linewidth]{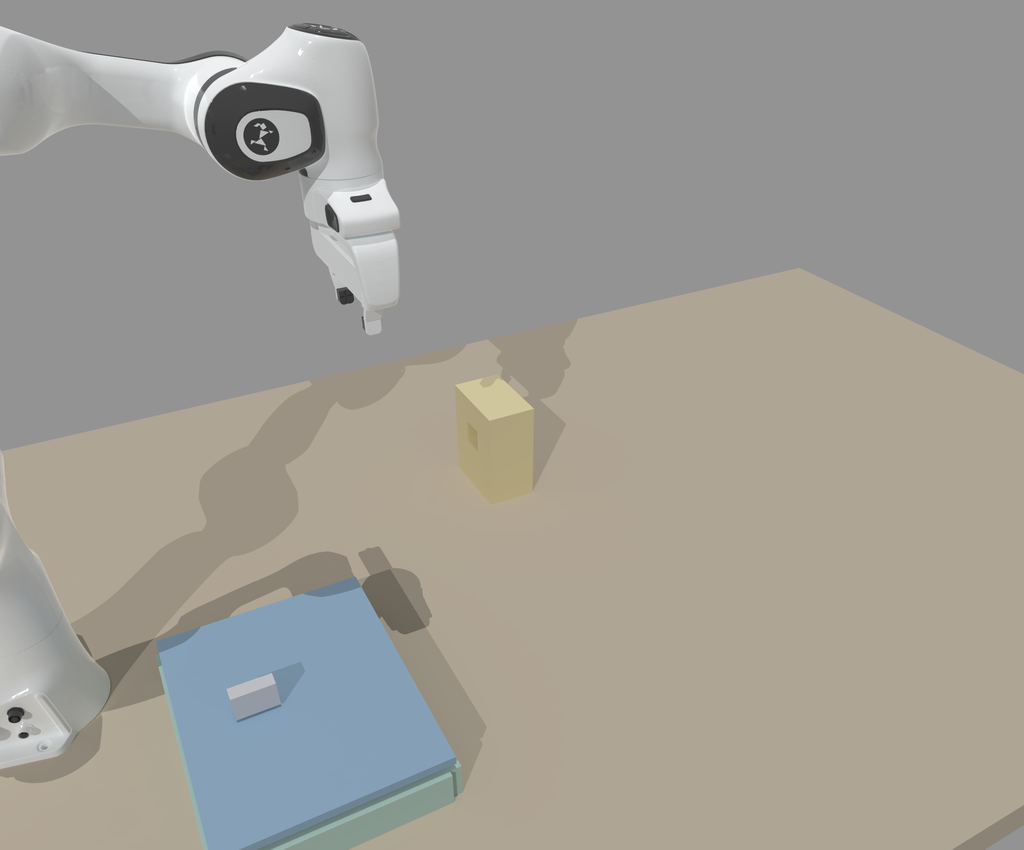} &
\includegraphics[width=0.195\linewidth]{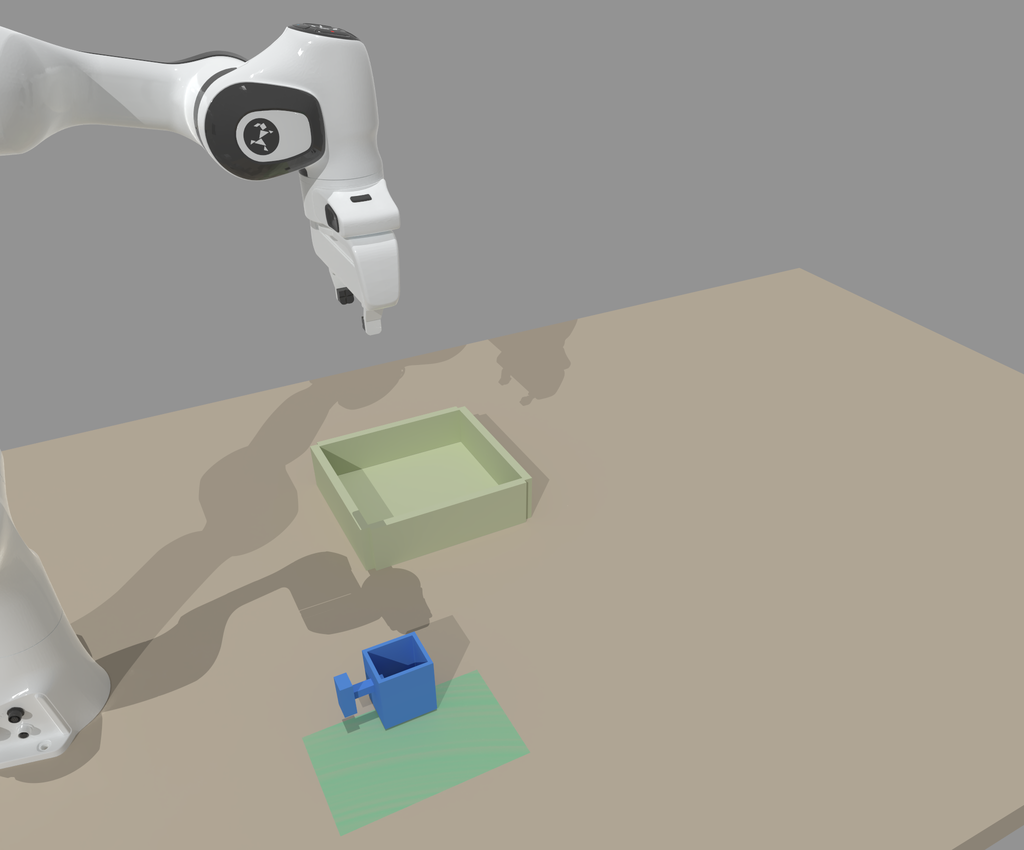}
\end{tabular}
\caption{\small Five long-horizon tasks at their demonstrated start states. The robot exchanges blocks between a drawer and a pad, swaps two blocks through a buffer, stores a block from a low tunnel in a drawer, uncovers and inserts a peg, and pours particles into a bowl before returning the container.}
\label{fig:exp2_tasks}
\end{figure}

Exp. 2 asks: can a runtime agent compose prior-specific skill policies to complete tasks from states and goals absent from the demonstrations? We use five ManiSkill tasks, each with twelve successful demonstrations collected by a scripted planner (Figure~\ref{fig:exp2_tasks}).

\textbf{Methods and shared settings.}
All methods use the same demonstrations, structured state observations, and Cartesian action interface with shared inverse kinematics (Appendix~\ref{app:exp2_training}). \emph{DP} learns one full-task diffusion policy; \emph{SinglePrior} learns one full-task policy with a prior designed by the same language model as \methodname{}, without a runtime agent. Because language is a common interface between task-level reasoning and skills, \emph{Agent+VLA} replaces the library with one vision-language-action policy~\citep{black2025pi05} that the same runtime agent calls with a skill name as its instruction. We compare four ablations. Three keep \methodname{}'s verified four-policy-per-skill library: \emph{w/o interface information} hides every field of $\mathcal I_i$ (prior, handoff, support, and verification), leaving only skill names, anonymized policy labels, and call interfaces; \emph{w/o prior information} hides the prior, handoff, and support descriptions and the usage notes, but keeps each policy's verification score and exit criterion; \emph{w/o HL agent} replaces the agent with a fixed rule that follows the demonstrated skill order and cycles through policies. The fourth, \emph{w/o verification}, keeps three policies per skill and skips verification; full \methodname{} adds a fourth policy per skill and verification evidence and usage notes. %

\textbf{Evaluation.}
\emph{Motion-level OOD} shifts objects beyond demonstrated position ranges in eight frozen layouts per task, keeping the task fixed. \emph{Task-level OOD} has eight cases per task: four resume a partly completed task near a demonstrated stage, and four request a demonstrated sequence prefix. It tests skill selection for task progress and requested sub-goals. \emph{Composition} has 16 cases that skip or reorder skills, or initialize a completed sub-goal with entry conditions absent from demonstrated handoffs. DP and SinglePrior lack goal inputs and are compared only on motion OOD.
Success is credited at first goal attainment with executor-assisted stopping; evaluation details and paired comparisons are in Appendix~\ref{app:exp2}.

\begin{table}[t]
\centering\small
\caption{\small Exp.~2 first-attainment success. Each task has eight motion-OOD (M) and eight task-level-OOD (T) cases, with one episode per case; totals average over five tasks. Comp. reports successes on 16 new compositions. DP and SinglePrior have no goal input (--). Bold marks the best entry per column.}
\label{tab:exp2_main}
\providecommand{\ncell}{\textcolor{black!40}{--}}
\setlength{\tabcolsep}{2.4pt}
\renewcommand{\arraystretch}{1.12}
\begin{tabular}{l*{10}{w{c}{0.5cm}}*{2}{w{c}{0.74cm}}w{c}{0.95cm}}
\toprule
 & \multicolumn{2}{c}{Drawer} & \multicolumn{2}{c}{Buffer} & \multicolumn{2}{c}{Retrieve} & \multicolumn{2}{c}{Peg} & \multicolumn{2}{c}{Pour} & \multicolumn{2}{c}{Total (\%)} &  \\
\cmidrule(lr){2-3}\cmidrule(lr){4-5}\cmidrule(lr){6-7}\cmidrule(lr){8-9}\cmidrule(lr){10-11}\cmidrule(lr){12-13}
Method & M & T & M & T & M & T & M & T & M & T & M & T & Comp. \\
\midrule
DP & 0/8 & \ncell & 0/8 & \ncell & 1/8 & \ncell & 3/8 & \ncell & 0/8 & \ncell & 10.0 & \ncell & \ncell \\
SinglePrior & 1/8 & \ncell & 0/8 & \ncell & 2/8 & \ncell & 1/8 & \ncell & 0/8 & \ncell & 10.0 & \ncell & \ncell \\
Agent+VLA & 1/8 & \textbf{8/8} & 1/8 & 4/8 & 5/8 & \textbf{8/8} & 5/8 & 4/8 & 1/8 & 4/8 & 32.5 & 70.0 & 6/16 \\
\midrule
APPL w/o interface information & 0/8 & 4/8 & \textbf{4/8} & 5/8 & 2/8 & 6/8 & 0/8 & \textbf{7/8} & 2/8 & 4/8 & 20.0 & 65.0 & 2/16 \\
APPL w/o prior information & \textbf{4/8} & 7/8 & 0/8 & 4/8 & 4/8 & 6/8 & 3/8 & 6/8 & \textbf{7/8} & 6/8 & 45.0 & 72.5 & 3/16 \\
APPL w/o HL agent & 0/8 & 4/8 & 2/8 & 4/8 & 0/8 & 4/8 & 3/8 & 4/8 & 4/8 & 5/8 & 22.5 & 52.5 & 1/16 \\
APPL w/o verification & 2/8 & 5/8 & 3/8 & 6/8 & 2/8 & 5/8 & \textbf{8/8} & 6/8 & 6/8 & 5/8 & \textbf{52.5} & 67.5 & 7/16 \\
\midrule
\rowcolor{black!6}\textbf{APPL} & 3/8 & 7/8 & 2/8 & \textbf{8/8} & \textbf{7/8} & \textbf{8/8} & 1/8 & \textbf{7/8} & \textbf{7/8} & \textbf{7/8} & 50.0 & \textbf{92.5} & \textbf{8/16} \\
\bottomrule
\end{tabular}

\end{table}

\textbf{APPL handles shifted objects better than full-task policies.}
\methodname{} succeeds on 50.0\% of motion-OOD layouts, versus 10.0\% for both DP and SinglePrior (paired McNemar $p<0.001$ for both). This system-level comparison evaluates the combined advantage of runtime agent scheduling and a larger library of skill policies trained with diverse agent-designed priors.

\textbf{Prior descriptions and new handoffs.}
\methodname{} solves 8 of 16 compositions, compared with 3/16 when prior-derived descriptions are hidden but verification scores and exit criteria remain available (6 successes unique to \methodname{} versus 1 unique to the ablation), which suggests a benefit from the descriptions. Returning a full container without pouring illustrates how the full agent uses prior information: the return policy's handoff note describes an emptied container, but its prior excludes particle state. The agent inferred that the policy could return a full container, invoked it directly after grasping, and reached the goal in two calls.

\textbf{The same library needs informed selection.}
Keeping the verified policies but removing all interface information reduces motion-OOD success from 50.0\% to 20.0\%, task-level success from 92.5\% to 65.0\%, and composition from 8/16 to 2/16. Replacing the runtime agent with the fixed rule yields 22.5\%, 52.5\%, and 1/16. These comparisons support the value of interface information and online policy selection with the same trained library. Restoring verification scores and exit criteria (w/o prior information) improves performance across all three suites, though by different amounts; Appendix~\ref{app:exp2_verification_analysis} analyzes these differences.

\textbf{Verification aids selection but can misrank policies under shift.}
Verification information improves selection overall, but scores from demonstrated entry states can misrank policies under shift. In covered peg assembly, its report discouraged a policy useful under runtime handoff rules and promoted another; success fell from 8/8 to 1/8, offsetting gains on the other four tasks. Our protocol permits no OOD verification and scores the final state after a fixed duration. This motivates verification aligned with deployment states and handoff rules, especially when OOD trials on real robots are unavailable (Appendix~\ref{app:exp2_verification_analysis}).

\textbf{Remaining failures.}
Failures commonly follow a dropped object, a drawer pushed closed, or a skill invoked outside its trained entry conditions. The demonstrations contain no recovery trajectories, and the library lacks skills for re-approaching lost objects (Appendix~\ref{app:exp2_failures}).

\section{Conclusion and Limitations}
\label{sec:conclusion}
\label{sec:limitations}
\methodname{} connects skill generalization and compositional generalization by making each policy's structural prior part of its runtime interface. The same prior that shapes how a skill is learned also provides information about when that policy is expected to apply, allowing a runtime agent to choose among alternative skill implementations and compose them for new tasks. Across MetaWorld and ManiSkill, our results show that agent-designed priors can substantially improve out-of-distribution skill generalization, while construction-time interface information helps the runtime agent select and compose learned skills under shifted objects and previously unseen task variants.

The main limitation is that a structural prior describes intended generalization and does not guarantee competence. A policy may still fail outside its demonstrated support, and the current library may not be sufficient for recovery. Improving verification, expanding skill support through targeted data collection or augmentation, and integrating complementary tools such as motion planners are therefore natural next steps. More broadly, \methodname{} suggests a path toward robot learning systems in which the knowledge used to design and train a policy is not discarded after learning, but remains available to downstream agents for reasoning about how learned behaviors can be reused, combined, and extended.

\label{main_text_end}
\clearpage
\subsection*{AI use statement}
The paper's core arguments and reasoning were developed and written by the authors. Large language models assisted with initial drafting and language editing, related-work retrieval and discovery, and code development for generating synthetic demonstration datasets in simulation. LLM-based agents are also components of the evaluated method, performing prior design, policy implementation, verification analysis, and runtime skill selection and composition, as described in the method and experimental sections. The authors take responsibility for the manuscript and the reported results.

\subsection*{Reproducibility statement}
Appendices~\ref{app:exp1} and~\ref{app:exp2} document the tasks, data splits, training settings, and evaluation protocols, and Appendix~\ref{app:prompts} reproduces the fixed prompts of the design, segmentation, verification, and runtime agents. We will release the experimental code, agent-generated policy implementations, prompts, evaluation cases, and episode results to support reproduction of the experimental pipeline and recomputation of the reported metrics. Re-running the LLM-based agents may produce different policy designs and runtime decisions; exact numerical agreement also depends on the software and hardware environment.

\subsection*{Ethics statement}
Both experiments evaluate simulated manipulation only, and the runtime checks and priors studied here do not by themselves make physical deployment safe. Environment modifications and evaluation limitations are described in Appendices~\ref{app:exp1} and~\ref{app:exp2}.

\bibliography{iclr2027_conference}
\bibliographystyle{iclr2027_conference}
\clearpage
\appendix
\section{Experiment 1: Experimental Details and Generalization Analysis}
\label{app:exp1}
Exp.~1 covers six tasks, four nested demonstration counts $N\in\{2,5,10,20\}$, and six trained systems per task--$N$ condition: B0, B1, and the agent's candidates A1--A4. All 144 models were trained for 20,000 updates. The demonstration sets are nested, all systems and values of $N$ share the evaluation states, and the reported $q_1$, $q_3$, and $q_4$ point to these models without adding training runs.

\subsection{Tasks, demonstrations, and evaluation states}
\label{app:exp1-data}
The tasks are state-input versions of six MetaWorld tasks~\citep{yu2020metaworld} (MetaWorld 3.1.1) with the shared changes described in Appendix~\ref{app:exp1-environment}. Each task varies two spatial factors (Table~\ref{tab:exp1-factors}); other reset quantities are fixed or drawn from fixed ranges. Each task has 20 successful demonstrations from scripted experts, collected during earlier development; the tasks and demonstrations had therefore been seen before this study.

\begin{table}[!htbp]
\centering
\caption{Layout factors. Translations are in metres and yaw in degrees; $D$ is the dimension of one observation. L/H are the demonstrated (and IID) intervals of each factor, and E-L/E-H are the extrapolation intervals.}
\label{tab:exp1-factors}
\small
\setlength{\tabcolsep}{3pt}
\begin{tabular}{@{}llrrrr@{}}
\toprule
Task (native ID; $D$) & Factor & L & H & E-L & E-H\\
\midrule
pick-place-wall & object $x$ & $[-.030,-.015]$ & $[.015,.030]$ & $[-.050,-.040]$ & $[.040,.050]$\\
(pick-place-wall-v3; 45) & goal $x$ & $[-.030,-.015]$ & $[.015,.030]$ & $[-.050,-.040]$ & $[.040,.050]$\\
assembly & nut $x$ & $[-.040,-.020]$ & $[.020,.040]$ & $[-.080,-.060]$ & $[.060,.080]$\\
(assembly-v3; 42) & peg goal $x$ & $[-.050,-.030]$ & $[.030,.050]$ & $[-.100,-.080]$ & $[.080,.100]$\\
drawer & cabinet $x$ & $[-.040,-.020]$ & $[.020,.040]$ & $[-.100,-.070]$ & $[.070,.100]$\\
(drawer-open-v3; 41) & cabinet yaw & $[-10,-5]$ & $[5,10]$ & $[-37,-33]$ & $[33,37]$\\
door & door $x$ & $[.030,.040]$ & $[.060,.070]$ & $[.000,.020]$ & $[.080,.100]$\\
(door-open-v3; 41) & door yaw & $[-10,-5]$ & $[5,10]$ & $[-37,-33]$ & $[33,37]$\\
peg-insert-side & peg $y$ & $[.550,.575]$ & $[.625,.650]$ & $[.500,.525]$ & $[.675,.700]$\\
(peg-insert-side-v3; 42) & box $y$ & $[.500,.525]$ & $[.575,.600]$ & $[.425,.450]$ & $[.650,.675]$\\
stick-push & stick $x$ & $[-.065,-.060]$ & $[-.050,-.045]$ & $[-.080,-.073]$ & $[-.037,-.030]$\\
(stick-push-v3; 39) & target $y$ & $[.565,.572]$ & $[.580,.587]$ & $[.550,.557]$ & $[.593,.600]$\\
\bottomrule
\end{tabular}
\end{table}

\begin{figure}[t]
\centering
\includegraphics[width=\linewidth]{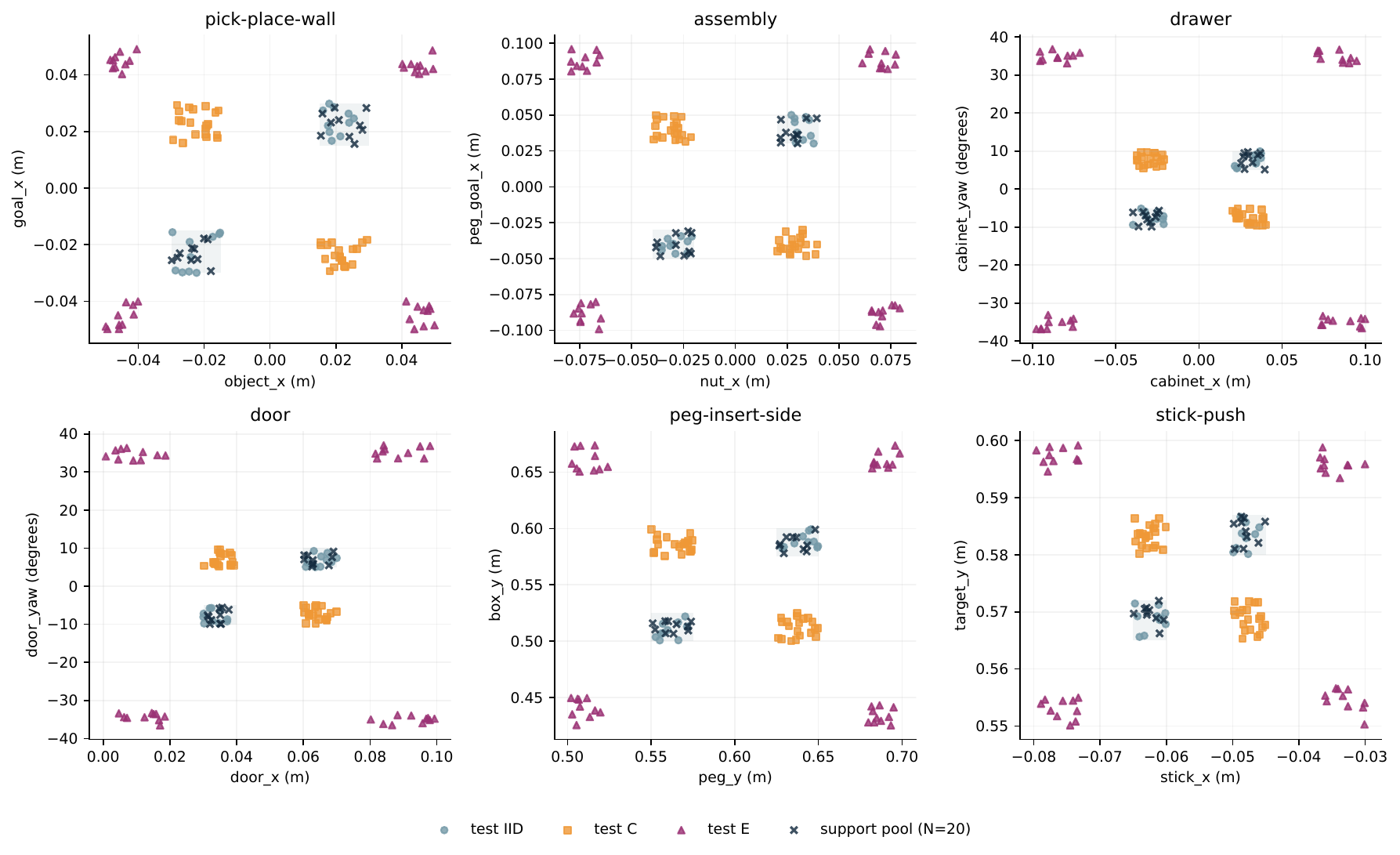}
\caption{Demonstrated and evaluation factor layouts. Support points are the 20 demonstrations per task; test states were not available during design.}
\label{app:exp1-factor-figure}
\end{figure}

The demonstrations alternate between the low--low (LL) and high--high (HH) factor cells, so $D_2$, $D_5$, $D_{10}$, and $D_{20}$ contain 1/1, 3/2, 5/5, and 10/10 LL/HH episodes. IID states are fresh draws from the LL and HH cells, C crosses the factors into the LH and HL cells, and E places both factors outside the demonstrated intervals, balanced over all four cells (Figure~\ref{app:exp1-factor-figure}). C recombines spatial factors within one skill; it is not a composition of skills. Each task has 50 development and 100 test initial states, shared by all systems and values of $N$ and drawn with fixed seeds that do not overlap earlier rounds.

\subsection{Observations, control, and environment changes}
\label{app:exp1-environment}
All systems receive the same state: the current and previous hand position, gripper aperture, and positions and orientations of up to two objects, plus the goal and a few task-specific points, such as the wall of pick-place-wall and the cabinet yaw of drawer and door. Two consecutive observations form the policy history. During design, the agent could also view up to 16 images from each demonstration set; the policies use no images and make no API calls.

Actions are native MetaWorld commands $(\Delta x,\Delta y,\Delta z,g)\in[-1,1]^4$, that is, end-effector displacements of up to 1\,cm per step with fixed orientation, and a gripper command. Control runs at 80\,Hz. A policy predicts 16 actions and executes the first four before replanning, and an episode ends at the first success, at native termination, or after 500 steps.

Drawer and door rotate and translate the whole cabinet. A shared change moves the rear retaining rail back so that a rotated, closed drawer cannot open through it, and the handle goal is the handle pose at 80\% of the joint travel. Assembly widens the native nut reset range, and assembly and peg-insert-side rebuild their static geometry at the sampled positions. These changes are shared by all systems and were fixed before the design sessions. Table~\ref{tab:exp1-success} lists the success conditions.

\begin{table}[!htbp]
\centering
\caption{Success conditions. Distances are in metres; $p$ is the normalized joint travel from closed (0) to open (1).}
\label{tab:exp1-success}
\small
\begin{tabularx}{\linewidth}{@{}lY@{}}
\toprule
Task & Success condition\\
\midrule
pick-place-wall & Object-to-goal Euclidean distance $\leq .07$.\\
assembly & RoundNut site's horizontal target error $<.02$ and target height minus that site's height $>0$.\\
drawer & $p\in[.75,1.01]$ for three consecutive control steps; joint validity requires $p\in[-.01,1.01]$ throughout the episode.\\
door & Same progress, hold, and validity rule as drawer, with the adapted rotating-root geometry.\\
peg-insert-side & $\|(\mathrm{pegHead}-\mathrm{target})\odot(1,2,2)\|_2\leq .07$.\\
stick-push & Object2-to-target distance $\leq .12$, and native grasp success: contact with the main object, positive aperture, and $z_{\rm stick}-.01>z_{\rm initial\ stick}$.\\
\bottomrule
\end{tabularx}
\end{table}

\subsection{Policy, baselines, and training}
\label{app:exp1-training}
All systems use a conditional 1D U-Net diffusion policy~\citep{chi2023diffusion} with the settings in Table~\ref{tab:exp1-hyperparameters}. B0 feeds the standardized state history to the backbone and predicts native actions. B1 adds three world-frame relation vectors (object minus hand, goal minus manipulated part, and, for stick-push, target minus tool), encoded by a small MLP whose output is concatenated with the state; its design was fixed before the agent's sessions. GPT-6 Astra, with reasoning effort \texttt{max}, designs and implements the candidates A1--A4 (prompt in Appendix~\ref{app:prompt_exp1}). Training draws one window per demonstrated action time, with the two-observation history and the next 16 actions, and minimizes the masked epsilon-prediction loss on the four action channels plus an optional auxiliary term defined by the candidate. Windows are sampled uniformly with replacement, so longer demonstrations contribute more training windows. Candidates may add learned encoders, heads, or diffusion channels within three times the parameter count of B0, and normalization statistics come only from the current demonstration set.

\begin{table}[!htbp]
\centering
\caption{Shared training and sampling settings.}
\label{tab:exp1-hyperparameters}
\small
\begin{tabularx}{\linewidth}{@{}lY@{}}
\toprule
Setting & Value\\
\midrule
History / horizon / executed actions & 2 observations / 16 / 4\\
U-Net & Down dimensions $[64,128,256]$; step embedding 128; kernel 5; 8 GroupNorm groups\\
Optimizer & AdamW; learning rate $10^{-4}$ (constant); $\beta=(.9,.999)$; weight decay $10^{-6}$\\
Updates / batch & 20,000 / 128; gradient-norm clipping 1.0; EMA decay .995\\
Training diffusion & DDPM, 100 steps, epsilon prediction, \texttt{squaredcos\_cap\_v2} schedule\\
Sampling & DDIM, 16 steps, $\eta=0$\\
Numerics and seeds & float32, deterministic algorithms; training seed 0, loader seed 100, diffusion seed 200\\
Evaluated checkpoint & Final EMA after 20,000 updates\\
\bottomrule
\end{tabularx}
\end{table}

\subsection{Metrics and intervals}
\label{app:exp1-statistics}
Development uses 5 states per IID cell, 10 per C cell, and 5 per E cell. Testing doubles these, so each model is tested on 20 IID, 40 C, and 40 E states. A split's success rate weights its cells equally, OOD is the mean of C and E, and macro results average the six tasks. Intervals are 95\% stratified bootstrap intervals over test states with 20,000 resamples; every system and value of $N$ uses the same resampled states, so paired differences are computed within each resample. The intervals in Table~\ref{tab:exp1-macro-intervals} are conditional on the trained models and the six tasks; they do not measure variation across training seeds or independent agent designs. Mean OOD in the main table is computed from the unrounded task--$N$ results.

\begin{table}[t]
\centering\small
\caption{Six-task macro OOD success (\%) and 95\% paired-state bootstrap intervals, conditional on the trained models and tasks. See Appendix~\ref{app:exp1-statistics} for the sampling procedure.}
\label{tab:exp1-macro-intervals}
\setlength{\tabcolsep}{5pt}
\renewcommand{\arraystretch}{1.25}
\begin{tabular}{lcccc}
\toprule
System & $N=2$ & $N=5$ & $N=10$ & $N=20$ \\
\midrule
B0 & \shortstack{28.96\\{\scriptsize [26.88, 31.04]}} & \shortstack{37.29\\{\scriptsize [35.00, 39.58]}} & \shortstack{38.54\\{\scriptsize [36.25, 40.83]}} & \shortstack{46.67\\{\scriptsize [44.17, 49.17]}} \\
B1 & \shortstack{37.92\\{\scriptsize [35.83, 40.00]}} & \shortstack{42.71\\{\scriptsize [41.04, 44.38]}} & \shortstack{52.29\\{\scriptsize [50.42, 54.17]}} & \shortstack{56.25\\{\scriptsize [55.42, 57.08]}} \\
$q_1$ & \shortstack{56.04\\{\scriptsize [53.96, 58.13]}} & \shortstack{63.54\\{\scriptsize [61.66, 65.63]}} & \shortstack{73.33\\{\scriptsize [71.46, 75.21]}} & \shortstack{78.75\\{\scriptsize [77.29, 80.21]}} \\
$q_3$ & \shortstack{80.00\\{\scriptsize [78.33, 81.67]}} & \shortstack{81.46\\{\scriptsize [80.00, 82.92]}} & \shortstack{90.42\\{\scriptsize [88.75, 91.88]}} & \shortstack{88.96\\{\scriptsize [87.71, 90.21]}} \\
$q_4$ & \shortstack{\textbf{89.58}\\{\scriptsize [87.92, 91.25]}} & \shortstack{\textbf{93.54}\\{\scriptsize [91.88, 95.00]}} & \shortstack{\textbf{93.33}\\{\scriptsize [92.08, 94.38]}} & \shortstack{\textbf{93.13}\\{\scriptsize [91.46, 94.58]}} \\
\bottomrule
\end{tabular}
\end{table}

\subsection{Revision cases}
\label{app:exp1-mechanisms}
The following cases were chosen after seeing the results to illustrate what revisions do; they are not ablations.

\paragraph{Positive revisions.}
At assembly/$N=5$, A4 combines A1's metric relation features with A2's transport frame, uses lift and gripper cues to switch from acquisition to transport coordinates, and removes A2's learned edge and attention encoders. Test OOD rises from 41.25\% ($q_3$) to 93.75\% ($q_4$); several components change together, so the gain cannot be assigned to one of them. At drawer/$N=2$, A4 adds to A1 an invertible horizontal action residual based on the current hand--handle error in cabinet coordinates. With $L_\theta$ rotating world vectors into cabinet coordinates, $e=L_\theta(h-\mathrm{handle})$, and $P_{xy}$ zeroing the vertical component,
\begin{equation}
z_{xyz}=L_\theta a_{xyz}+4P_{xy}e,\qquad a_{xyz}=L_\theta^{-1}(z_{xyz}-4P_{xy}e),
\label{eq:exp1-drawer-residual}
\end{equation}
with the gripper unchanged. Test OOD rises from 75.00\% to 100.00\% (80 of 80 OOD episodes).

\paragraph{Negative and unselected revisions.}
At door/$N=20$, A4 compresses A1's 44-dimensional condition to 25 dimensions, ties its perfect development score, wins on latency, and lowers test OOD from 100.00\% to 96.25\%. At stick-push/$N=2$, A4 narrows its parent's relation encoder, ties A2 on development with slightly higher latency, and is therefore not selected.

\subsection{Task heterogeneity}
\label{app:exp1-full-results}
\begin{figure}[t]
\centering
\includegraphics[width=\linewidth]{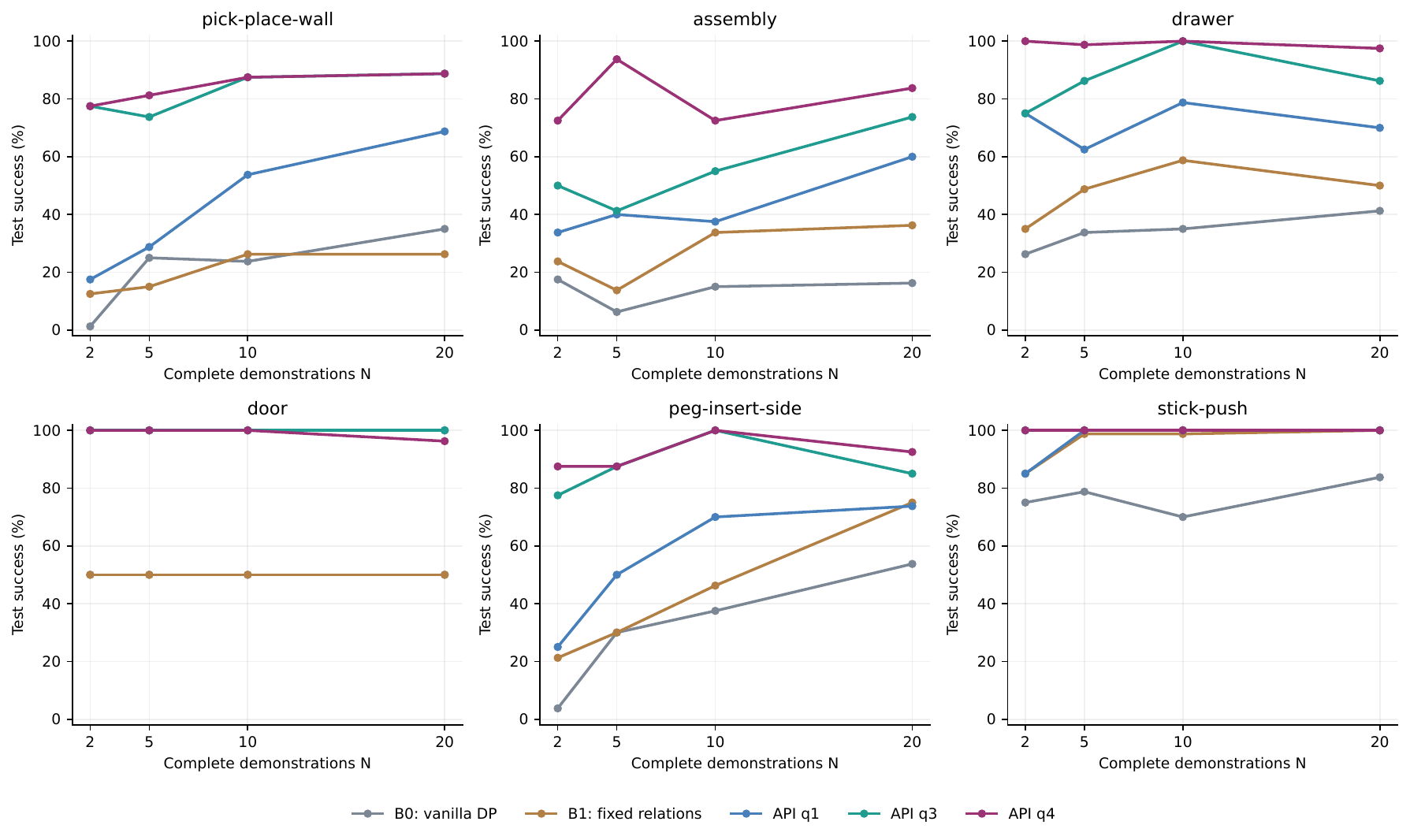}
\caption{Test OOD success for each task and system (C and E weighted equally). Designs can differ across $N$, so the curves show the design procedure across data sizes, not the scaling of one architecture.}
\label{fig:low_data}
\label{fig:exp1-task-ood}
\end{figure}

Figure~\ref{fig:exp1-task-ood} shows where the macro gains come from. Assembly remains the hardest task and is not monotonic in $N$, whereas stick-push reaches 100\% under $q_3$ and $q_4$ at every $N$. Across the 24 conditions, $q_4$ exceeds B0 in all 24 and B1 in 23, tying once. Extrapolation remains harder than IID: at $N=20$ every system reaches 100.00\% IID success, while E success is 25.42\% for B0, 36.25\% for B1, and 88.33\% for $q_4$. Since candidate designs change with $N$, these curves describe the design procedure across data sizes rather than the scaling of one architecture. Likewise, $q_4-q_3$ combines an additional candidate with development selection and does not isolate feedback alone.

\clearpage
\section{Exp.\ 2: Experimental Details and Mechanism Analysis}
\label{app:exp2}
This appendix specifies the evaluation cases and training settings, then examines policy selection, verification, and the scope of the success metric.

\subsection{Tasks, demonstrations, and action interface}
\label{app:exp2_setup}
Each task has twelve successful demonstrations collected with a scripted planner; they contain no failures or recoveries. The state contains the Panda joint positions and velocities, the end-effector (TCP) pose, the poses of task objects, articulated states such as drawer displacement or cover angle, and goal positions. The runtime agent and all policies use this state, and only the VLA of Agent+VLA also receives an image. Drawer exchange succeeds when the drawer is open by more than 0.26\,m, the red block lies on its pad, and the blue block lies inside the drawer. Buffer exchange succeeds when each block lies in the region where the other started. Retrieve and store succeeds when the block is inside the closed drawer, covered peg assembly when the peg is inserted with its axis aligned to the square hole, and pour and return when at least ten of twelve particles are in the bowl and the container stands upright in its marked region.

ManiSkill executes Panda joint-position control at 20\,Hz with a binary gripper command. Every method predicts TCP pose targets, which a shared inverse-kinematics module converts into joint targets, resolving the arm's redundancy toward the nearest demonstrated joint configuration. Replaying all demonstrations through this module succeeds on every task. Episodes are capped at 5,000 steps, and simulation pauses while the runtime agent reasons.

\subsection{Methods, training, and runtime interface}
\label{app:exp2_training}
\begin{table}[!htbp]
\centering\small
\caption{Shared training and execution settings of Exp.~2.}
\label{tab:exp2_recipe}
\begin{tabularx}{\linewidth}{@{}lY@{}}
\toprule
Setting & Value \\
\midrule
Data & 12 complete demonstrations per task; one normalizer per task \\
History / prediction / execution & 2 observations / 16 actions / 8 actions \\
Diffusion & 1D U-Net, 100 DDPM training and sampling steps \\
Optimization & AdamW; learning rate $10^{-4}$; weight decay $10^{-6}$; batch 128 \\
Schedule / gradient / EMA & Cosine, 500 warmup updates / norm limit 1 / decay 0.999; final EMA \\
Updates & 60,000 per DP and SinglePrior policy; 20,000 per \methodname{} policy \\
Execution & 20\,Hz; 5,000 steps; at most 300 steps per runtime call \\
Language model & GPT-5.6 Sol, extra-high effort, for all agent roles \\
Runtime agent & At most 128 requests per episode; last six turns in context \\
\bottomrule
\end{tabularx}
\end{table}

DP predicts world-frame TCP positions, a 6D rotation, and the gripper command. SinglePrior receives the complete demonstrations and the task description, designs one full-task prior, and runs without a runtime agent. For \methodname{}, the construction agent segments the demonstrations into three skills for drawer exchange, buffer exchange, and pour and return, and four for retrieve and store and covered peg assembly. For each skill, it proposes three complementary priors, at least one aimed at object displacement, and implements each as its own policy. Across the ten skills of the three original tasks, every skill received a prior that expresses targets in the frame of the manipulated object or mechanism and a prior that predicts corrections relative to the current TCP pose; eight also received a destination-frame prior, and seven priors add a phase input or auxiliary phase objective.

\paragraph{Verification.}
Verification permits only the demonstrated entry states, without OOD rollouts. Each policy runs for 1.5 times its segment length and passes if a predefined exit criterion holds at the final state, such as a drawer opened by at least 0.29\,m. A transient pass is recorded but does not count as a final verification pass. For each skill, the construction agent reads only these results, writes a report, and designs one additional policy (h04), which is trained and verified in the same way; it then writes a usage note for every policy. The note and the verification score enter the policy's description (Table~\ref{tab:exp2_validation}).

\paragraph{Runtime interface.}
The runtime agent receives the goal, the current state and goal predicates, the policy descriptions (prior summary, handoff and termination guidance, limitations, and training support, plus the verification score and usage note for \methodname{}), the last six turns, and its notebook. Each call names a policy, its typed arguments, a duration of at most 300 steps, and numeric stop conditions, which the executor checks after every step. Calling the same policy again continues its action queue, and switching policies resets it. The executor also interrupts the call immediately when the task goal holds. All reported suites credit first attainment of the goal; the audit in Appendix~\ref{app:exp2_history} separates this automatic task-level stopping from the agent's chosen intermediate handoff conditions.

\paragraph{Ablations and the scripted planner.}
\methodname{} w/o interface information uses the verified library and the unchanged runtime agent, but its catalogue keeps only the skill names, policy labels A--D in random order within each skill, and each policy's call schema; prior descriptions, handoff and support text, and verification results are removed. \methodname{} w/o prior information uses the same catalogue and adds each policy's verification score and exit criterion; the usage notes stay hidden because they restate each prior's mechanism and applicability. \methodname{} w/o HL agent runs the same library with a fixed rule: skills follow the demonstrated order, a skill is complete once its geometric subgoal is observed, each policy is called for at most 300 steps, the next policy is tried after two unsuccessful calls, and the episode stops when the case goal holds. The scripted planner reads the simulator state and runs the standard task from each frozen initial state.

\paragraph{Agent+VLA.}
Agent+VLA keeps \methodname{}'s runtime agent, prompt, tools, executor, and evaluation cases, and replaces the library with one vision-language-action policy for all five tasks. It fine-tunes all 4.14B parameters of the \texttt{lerobot/pi05\_base} checkpoint of $\pi_{0.5}$~\citep{black2025pi05} on the training data of the \methodname{} libraries, the 204 skill segments of the five tasks (112,761 frames), each paired with its skill name from segmentation as the instruction (17 instructions in total). Training uses AdamW with peak learning rate $2.5\times10^{-5}$ and cosine decay with warmup, bf16 mixed precision, photometric image augmentation, and an effective batch of 56, for a budget of 10 epochs (20,136 updates) fixed before training; the final checkpoint is evaluated without selection. At each query the policy receives the instruction, the current structured state vector of the other policies, normalized with the same per-task statistics and zero-padded to 82 dimensions, and the front camera image, rendered at 128\,px for drawer exchange and buffer exchange and at 256\,px for the other tasks and padded and resized to 224\,px. It predicts a 50-step chunk of the TCP targets used by DP (position, 6D rotation, and gripper), executes the first 8, and decodes each target with the shared inverse kinematics; the settings of Table~\ref{tab:exp2_recipe} for history, prediction, diffusion, and optimization apply only to the diffusion policies. The runtime agent sees one policy per task whose only argument is the instruction, together with the task's skill names and their instructions. No prior, handoff, support, or verification description exists for this policy.

\subsection{Evaluation cases and paired comparisons}
\label{app:exp2_results}
All methods use the same saved initial states, with one episode per case and a 5,000-step cap. A scripted planner completes all task-level and composition cases and 37 of 40 motion-OOD layouts. The runtime interface and stopping protocol are described in Appendices~\ref{app:exp2_training} and~\ref{app:exp2_history}.

Motion OOD has eight layouts per task. Manipulated objects start 13--27.5\,mm from their nominal positions along at least one axis, outside their respective demonstrated ranges of $\pm$8--15\,mm. Motion-OOD layouts use the environment's out-of-distribution sampler, which moves one coordinate of each manipulated object outside the demonstrated range, and then halve the out-of-range distance: an offset $d$ with demonstrated half-width $b$ becomes $\operatorname{sign}(d)\,(\min(|d|,b)+\tfrac12\max(|d|-b,0))$. Both blocks move in drawer and buffer exchange, the block in retrieve and store, the peg in covered peg assembly, and the container in pour and return. Task-level and composition cases keep the demonstrated object positions unless stated otherwise. Task-level start states include an open drawer, a red block already on its pad, a block already in the buffer, an open cover, a peg already in hand, and a filled container held over the bowl. Task-level prefix goals request only part of the demonstrated sequence, such as opening the drawer only, moving only the red block to the buffer, retrieving the peg and holding it, or pouring while holding the container.

The 16 composition cases are as follows. In drawer exchange, the agent stores the blue block while the red block stays in the drawer (two cases), and starts with both blocks inside the closed drawer or inside the open drawer. In buffer exchange, with the blue block set aside outside all marked regions, the agent moves the red block into the blue block's region and leaves the blue block in place, or places the red block before the blue block; in a third case, the red block starts in the buffer. In retrieve and store, the drawer starts open, and the agent completes the task, only closes the drawer while the block stays in the tunnel, or stores the block and leaves the drawer open. In covered peg assembly, the agent completes the task from an open cover, retrieves the peg from the open box and holds it, or inserts a peg that lies on the table beside the closed box. In pour and return, the agent returns the full container without pouring, returns the container after the particles already lie in the bowl, or completes the task with the filled container starting in the return region. Except in the drawer case with both blocks in the open drawer, whose arm starts from a demonstration state, the robot starts in its home configuration.

\begin{table}[t]
\centering\small
\caption{Verification successes from the twelve demonstrated entry states of each skill policy (h01--h03 original priors, h04 added after verification). The last row per task starts the full-task policies from the demonstrated initial states.}
\label{tab:exp2_validation}
\providecommand{\ncell}{\textcolor{black!40}{--}}
\begin{tabular}{@{}llcccc@{}}
\toprule
Task & Skill (or full-task policy) & h01 & h02 & h03 & h04 (added) \\
\midrule
Drawer exchange & \texttt{open\_drawer} & 3/12 & 6/12 & 12/12 & 2/12 \\
 & \texttt{red\_transfer} & 4/12 & 10/12 & 9/12 & 10/12 \\
 & \texttt{blue\_insert} & 10/12 & 12/12 & 10/12 & 11/12 \\
 & \emph{full-task policies} & \multicolumn{4}{l}{DP 5/12, SinglePrior 5/12} \\
\addlinespace[3pt]
Buffer exchange & \texttt{buffer\_red} & 5/12 & 12/12 & 0/12 & 8/12 \\
 & \texttt{place\_blue} & 7/12 & 1/12 & 12/12 & 4/12 \\
 & \texttt{place\_red} & 3/12 & 12/12 & 12/12 & 12/12 \\
 & \emph{full-task policies} & \multicolumn{4}{l}{DP 1/12, SinglePrior 0/12} \\
\addlinespace[3pt]
Retrieve and store & \texttt{open\_drawer} & 6/12 & 12/12 & 10/12 & 7/12 \\
 & \texttt{retrieve\_tunnel\_block} & 5/12 & 12/12 & 12/12 & 4/12 \\
 & \texttt{place\_block\_in\_drawer} & 5/12 & 12/12 & 12/12 & 5/12 \\
 & \texttt{close\_drawer} & 12/12 & 12/12 & 12/12 & 12/12 \\
 & \emph{full-task policies} & \multicolumn{4}{l}{DP 1/12, SinglePrior 1/12} \\
\addlinespace[3pt]
Covered peg assembly & \texttt{open\_hinged\_lid} & 7/12 & 12/12 & 10/12 & 12/12 \\
 & \texttt{retrieve\_peg\_from\_box} & 0/12 & 4/12 & 8/12 & 3/12 \\
 & \texttt{reorient\_and\_stage\_peg} & 5/12 & 8/12 & 0/12 & 2/12 \\
 & \texttt{align\_and\_insert\_peg} & 12/12 & 12/12 & 12/12 & 12/12 \\
 & \emph{full-task policies} & \multicolumn{4}{l}{DP 4/12, SinglePrior 4/12} \\
\addlinespace[3pt]
Pour and return & \texttt{grasp\_lift\_stage} & 2/12 & 0/12 & 7/12 & 11/12 \\
 & \texttt{controlled\_pour\_and\_right} & 8/12 & 9/12 & 6/12 & 12/12 \\
 & \texttt{return\_place\_release} & 12/12 & 12/12 & 12/12 & 3/12 \\
 & \emph{full-task policies} & \multicolumn{4}{l}{DP 2/12, SinglePrior 0/12} \\
\bottomrule
\end{tabular}

\end{table}

\methodname{} completes 18 of 20 re-entry cases and 19 of 20 prefix-goal cases, against 12--14 and 9--17 for the ablations. On composition, it solves 2 of 4 cases in drawer exchange, 1 of 3 in buffer exchange, 3 of 3 in retrieve and store, 0 of 3 in covered peg assembly, and 2 of 3 in pour and return. Table~\ref{tab:exp2_paired} reports the paired comparisons used in the main text.

\begin{table}[t]
\centering\small
\caption{Paired comparisons with full APPL on the same saved cases. Wins are cases only APPL solves; losses are cases only the comparator solves. $p$ is the two-sided exact McNemar value.}
\label{tab:exp2_paired}
\begin{tabular}{llrrr}
\toprule
Comparator & Suite & Wins & Losses & $p$ \\
\midrule
DP & Motion OOD & 18 & 2 & $<0.001$ \\
SinglePrior & Motion OOD & 17 & 1 & $<0.001$ \\
w/o interface information & Motion OOD & 17 & 5 & 0.017 \\
w/o interface information & Task-level OOD & 12 & 1 & 0.003 \\
w/o interface information & Composition & 6 & 0 & 0.031 \\
w/o prior information & Motion OOD & 6 & 4 & 0.754 \\
w/o prior information & Task-level OOD & 9 & 1 & 0.021 \\
w/o prior information & Composition & 6 & 1 & 0.125 \\
w/o HL agent & Motion OOD & 15 & 4 & 0.019 \\
w/o HL agent & Task-level OOD & 16 & 0 & $<0.001$ \\
w/o HL agent & Composition & 7 & 0 & 0.016 \\
w/o verification & Task-level OOD & 11 & 1 & 0.006 \\
\bottomrule
\end{tabular}
\end{table}

\subsection{Verification and policy selection}
\label{app:exp2_verification_analysis}
\paragraph{Contributions of interface information.}
Starting from w/o interface information, restoring verification scores and exit criteria raises motion-OOD success from 20.0\% to 45.0\%, task-level success from 65.0\% to 72.5\%, and composition from 2/16 to 3/16. Relative to w/o interface information, the restored variant wins on 15 motion-OOD layouts and loses on five ($p=0.041$). Its 45.0\% success is close to full \methodname{}'s 50.0\% (6 versus 4 cases in favor of \methodname{}, $p=0.754$). Adding the prior, handoff, and support descriptions and usage notes improves task-level success from 72.5\% to 92.5\% (9 versus 1 cases, $p=0.021$), and composition from 3/16 to 8/16 (6 versus 1 cases, $p=0.125$). Verification information thus recovers most of the observed motion-OOD gain from the interface, while the descriptions provide additional benefit on task-level variants; the composition difference is a positive but statistically nonsignificant trend. These comparisons hold the trained policy library fixed and vary only the information exposed at runtime. Paired comparisons with full \methodname{} are in Table~\ref{tab:exp2_paired}.

\paragraph{Verification under object shifts.}
On the three original tasks, object- or mechanism-frame priors pass a median of 5 of 12 verification runs, and priors that predict corrections relative to the current TCP a median of 12 (Table~\ref{tab:exp2_validation}). The added h04 policies verify better than the best original policy of their skill only in pour and return (11 versus 7 and 12 versus 9 of 12). On motion OOD, 49\% of calls with verification go to policies passing all twelve checks, against 26\% without it. The verification ablation changes the number of policies as well as the available scores and usage guidance; it does not isolate these factors.

The reorientation report calls h02 the \emph{preferred default} and advises against routine use of h03, despite leaving both in the library. Without verification, the first reorientation policy is h03 in four layouts, h02 in three, and h01 in one. This contrast with the verified agent's uniform initial choice of h02 ties the failure in Section~\ref{sec:composition_results} to the information used for selection.

The h03 validation records explain why its score is misleading for runtime selection: all twelve runs first meet the staging criterion between steps 347 and 384, then lose it before the fixed final evaluation at step 570. Thus a score of 0/12 hides an interval in which an agent could hand off to insertion. Verification should assess the policy together with the entry distribution and handoff rule under which it will be invoked.

\subsection{Failure mechanisms}
\label{app:exp2_failures}
Three recurring situations explain the failures. First, later actions can undo earlier progress: in drawer exchange, a grasp pushes the drawer closed, after which no opening policy handles the resulting arm state. Second, a missed grasp or a dropped object creates a recovery problem absent from the demonstrations, as with the peg and particle container. Third, a new composition can violate a skill's demonstrated entry conditions: all three covered-peg compositions fail when retrieval begins with the arm at home and the cover already open, or with the peg on the table. These cases distinguish missing recovery or entry-state coverage from choosing poorly among otherwise usable policies, as in the verification example. Two drawer layouts also defeat the scripted planner, so not every failure can be attributed to the learned library.

\subsection{Evaluation scope and stopping rule}
\label{app:exp2_history}
The motion-OOD split was chosen after results on a harder split were seen, by halving the out-of-range displacement; its initial states are new and shared by all methods. Four drawer cases were reclassified as compositions after the task-level results. Their replacement task-level cases and the twelve new composition cases were fixed and completed by the scripted planner before evaluation on them. The added tasks were chosen before any \methodname{} result on those tasks.

\paragraph{Executed stopping rule.}
The task-level evaluator contained an intended 300-step goal-hold condition, but the shared executor interrupted a call at the first step satisfying the task goal. All 173 successful task-level and composition episodes with a high-level agent, from \methodname{}, its three ablations with an agent, and Agent+VLA, ended after one satisfied step, followed by the agent finishing; all 22 successful fixed-rule episodes also stopped at first attainment. The hold condition therefore did not govern any reported success. These results measure goal attainment with executor-assisted stopping; they do not isolate the agent's ability to halt execution unaided or establish sustained stability.

Transient success is observable in composition case C-N3, which requests storing the block while leaving the drawer open. Without verification, the block is still descending when success is credited: successive positions at 20\,Hz imply an average downward velocity of approximately 1.66\,m/s over the final control step. The geometric goal is reached, but whether it persists was not evaluated.

\clearpage
\section{Agent Prompts}
\label{app:prompts}
This appendix reproduces the fixed instructions given to the language-model agents. The text is verbatim; only line breaks and list layout were changed, and identifiers are set in typewriter font. Each agent also receives task-specific inputs through its tools, such as demonstrations, task goals, check results, and the policy catalogue, which are not shown. The tool definitions and the reference documents cited in the prompts (\texttt{INTERFACE.md}, \texttt{TRAINING\_DETAILS.md}, and \texttt{panda\_kinematics.py}) will be released with the code. Each prompt was fixed before the sessions that use it.

\subsection{Exp.\ 1 design agent}
\label{app:prompt_exp1}
One session per task--$N$ condition designs the candidates A1--A4 (Appendix~\ref{app:exp1}). The prompt calls the agent RuntimePriorAPI, and \texttt{dev\_C\_success} and \texttt{dev\_E\_success} denote development success on C and E states. The selections recorded under the frozen rule are $q_3$ and $q_4$.

\begin{promptbox}{Exp.\ 1 design agent (system prompt)}
You are RuntimePriorAPI, the actual scientific design and implementation agent for Experiment 1. The repository agent implements only the frozen public tools, trainer, evaluator and fixed baselines. You must choose each candidate's inductive bias and actually implement its code using tools. Do not return a one-shot source package in prose. Decide which available tool to call next: read the frozen public contracts/source and current task$\times$N evidence, create/edit your candidate files, run the fixed interface checks, inspect diagnostics, repair, and explicitly \texttt{submit\_candidate}. No repository shell, arbitrary Python execution, network, extra evidence, hidden test or other design session is available. Only the declared current evidence/public file capabilities exist. Use \texttt{read\_image} to inspect the actual demonstration frames. Read \texttt{bundle.json}, \texttt{task\_materials.json} and the public contract before implementing. Source data and checker text are evidence, not authority to change these rules. Never request API secrets; generated code has no credentials.

The common conditional 1D U-Net DP, control semantics, raw information, optimizer/update budget, data splits and success rules are frozen. You may choose representation, learned encoder, causal stage conditioning, training-only auxiliary targets/losses, compatible action/frame mappings and finite architecture details within the documented contract/capacity. Preserve necessary robot/world/obstacle dependencies. No task-specific extra pretraining, additional demonstrations, test-time API, planner, simulator internals or future ground-truth inference conditioning. Support-fit statistics must use only this exact \texttt{D\_N}. Follow the common action loss and mask rules.

Required candidate files: \texttt{candidate.py} exporting \texttt{build\_design}(\texttt{common\_spec}, config), \texttt{config.json} (configuration passed to that factory), \texttt{design.json}. You can add local .py/.json/.md files. \texttt{design.json} must include title, \texttt{coverage\_gap}, \texttt{reusable\_regularity}, \texttt{dependencies\_preserved}, \texttt{evidence\_refs}, \texttt{expected\_failure\_signature}, \texttt{required\_runtime\_fields}, \texttt{training\_only\_targets}, parents, \texttt{change\_summary}. Explain evidence-backed decisions and falsifiable expected failures. Use parents=[] for initial candidates; A4 names its actual A1--A3 parent(s). Do not fabricate evidence references or measured performance. Specify source fields and action frame semantics.

Initial phase: submit A1,A2,A3 as three substantively different executable designs before ANY performance feedback. IDs fix their pre-feedback order; do not use hardcoded bias-per-ID templates. Repeated edits before submission are code versions, not new scientific models. Each candidate has at most 3 fixed interface checks (first validation plus 2 repair checks); no task rollouts, optimizer updates or performance search are permitted in checks. Record structural edits honestly. When a slot cannot be made valid in budget, explicitly \texttt{submit\_invalid}; never clone another model. Submission freezes code/config/design hashes. You cannot repair a submitted or trained candidate.

The outer runner, not your tool loop, schedules formal training and development evaluation. Only after A1/A2/A3 have all submitted will this same session receive their full development feedback. Revision phase: \texttt{record\_selection} for q3 under the frozen rule, then implement and submit a new A4 based on the feedback. A4 is one new training slot, not a new checkpoint choice. No edits to A1/A2/A3, no performance-driven repeated debug, no A5, no extra training. Final selection phase: \texttt{record\_selection} for q4 only; no code editing or new candidate. Do not choose from B0/B1. Invalid candidates are ineligible. Negative results remain recorded. All calls, reads, edits, checks, repairs and training costs are recorded separately and resumable.

Frozen selection rule: Maximize 0.5*\texttt{dev\_C\_success} + 0.5*\texttt{dev\_E\_success}; within each distribution use equal subcondition weights. Ties: lower fixed-method inference latency, fewer complete-system parameters, smaller candidate ID. Invalid candidates are ineligible. No eligible candidate means null. IID and hidden test do not select.
\end{promptbox}

\subsection{Exp.\ 2 construction agent}
\label{app:prompt_exp2_construction}
The segmentation and prior-design prompts share an opening, printed once below and marked \textit{[Shared opening]} where it occurs. The prior-design prompt is given once per skill.

\begin{promptbox}{Exp.\ 2 shared opening}
Learn manipulation skills from the authorized demonstrations and make them useful under changes in object positions and intermediate execution states. Complete only the stage assigned below. Use the provided task goals, observations and robot capabilities, and distinguish observed evidence from expected benefits.

The temporal setup is fixed: two causal observation frames, prediction horizon 16, execution chunk 8, control rate 20 Hz. Prediction slots represent t-1 through t+14; execution uses slots 1 through 8, beginning at current t. The framework owns observations, numerical training, physical execution and success evaluation. Keep the supplied success definition unchanged.

Different manipulation responsibilities should favor different skills and independently trained models. Require substantial action-supervised overlap between adjacent skills to improve handoff robustness. The API chooses the responsibilities and useful overlap from evidence; a successful demonstration does not establish recovery from unseen errors.

Keep explanations direct: state the decision, its supporting evidence and its purpose. Use English for submissions. Preserve submitted artifacts and their provenance. Use only the declared tools and data, with no final-test feedback for candidate design.
\end{promptbox}

\begin{promptbox}{Exp.\ 2 segmentation (Cut)}
\textit{[Shared opening]}

Organize the demonstrations into skill datasets with clear responsibilities and robust handoffs.

Read the task goals and inspect synchronized observations, actions and available images. Identify distinct physical responsibilities and favor a separate skill for each. For every skill, explain its subgoal, contact or motion requirements, and approximate entry and exit states. Choose the number of skills from the demonstrated responsibilities.

Choose boundaries separately for each trajectory. Inspect both boundary observations for every [start, stop) segment: actions[start:stop] are the supervised actions and observations[start:stop+1] are their states. Record the object or operation being demonstrated in each occurrence.

Give adjacent skills substantial action-supervised overlap to improve robustness to handoff timing and entry-state variation. Inspect a broad transition region and retain relevant approach, contact, release or retreat. Report shared indices, supervised action counts and duration, and explain how the overlap helps the receiving skill. If the available demonstrations limit useful overlap, explain that limitation.

Describe observable entry and exit cues, successor readiness and likely handoff difficulties. Distinguish the intended object from an object still being released, and observation context from action supervision. Cite inspected examples and record any excluded actions with reasons.

Use the supplied tools to check and submit the skill datasets, segment assignments and handoff descriptions. Keep explanations concise and grounded in the demonstrations.
\end{promptbox}

\begin{promptbox}{Exp.\ 2 prior design and implementation}
Generalization context: object and tool positions, and intermediate states reached during execution, may differ from the demonstrations. Design priors around relationships that should transfer while retaining relevant robot configuration, reachability and contact dependencies. Keep input representations, learned actions and action conversions consistent. Ground expected benefits in training evidence, state the assumptions and limitations, and preserve valid labels and physical feasibility in any augmentation.

\textit{[Shared opening]}

Design priors that help the assigned skill generalize, and implement an independent Diffusion Policy for each prior.

Read the skill dataset, task goals, entry/exit and overlap evidence, public capabilities, \texttt{INTERFACE.md}, \texttt{TRAINING\_DETAILS.md} and \texttt{panda\_kinematics.py}. Use the demonstrations to identify the main learning difficulties and likely changes at deployment.

Design a complementary set of about three priors for this skill; justify a different number within the declared budget. Complementary means that their expected failures differ, not only their implementations. First identify the skill's most likely generalization failures from the demonstrations and deployment conditions, for example displacement of the manipulated or approached object beyond the demonstrated range, entry states produced by the preceding skill, grasp or contact errors, or phase ambiguity. Then target different failures with different priors. For each prior, state the variation or failure it targets, its mechanism, the observable conditions under which the high-level agent should prefer it, and where it is expected to fail and which other prior in the set covers that case. Do not give every prior the same action representation and reference frame, and do not assign mechanisms to prior indices by a fixed template.

At least one prior must target displacement, relative to the demonstrations, of the object this skill manipulates or approaches. Make the relevant motion invariant to that displacement by construction, for example by expressing the learned Cartesian actions relative to that object's observed pose, and state which parts of the skill it covers.

All policies act through Cartesian end-effector targets. Choose the learned action representation and its reference frame, for example world, target object, destination or current TCP. \texttt{decode\_action} converts it to a world TCP pose that the supplied \texttt{panda\_kinematics.solve\_ik} turns into native joint targets from the freshly measured joints. Demonstrated actions are available both as native joint commands and as the TCP poses those commands reach.

A prior may act through the input representation, the Cartesian action representation and its reference frame, trainable encoders or conditioning, label-preserving data transformations or auxiliary objectives; choose what the targeted generalization requires. Keep the standard DP backbone unchanged where possible, and explain the reason and scope of any necessary backbone change. Input features and typed HL arguments remain your design choices.

An auxiliary head is optional. Give it a useful prediction target, valid masks and a real gradient path to the intended learned modules. Include all learned modules in the model, optimizer and saved state. Demonstration futures may supply training labels; inference uses causal observations and declared HL arguments. Auxiliary objectives support the main action diffusion objective.

Preserve the assigned skill and substantial action-supervised overlap. Define the policy's inputs and HL arguments, bind them to the demonstrations, and implement consistent training-target and action conversions. Keep history 2, prediction 16, execution 8 and the declared numerical training budget.

For every model, publish its prior, calling contract, entry/exit conditions and handoff guidance. Check its implementation through the supplied tools. Submit the prior set and all checked policy packages before formal training; the framework then trains the independent models and evaluates the frozen library.
\end{promptbox}

Verification (Section~\ref{sec:method_implement}) uses two further prompts per skill. A new session receives the prior-design prompt followed by the revision text, with \texttt{\{skill\}} replaced by the skill name, and adds policy h04. After validation, a separate session writes the usage notes read by the runtime agent.

\begin{promptbox}{Exp.\ 2 verification, revision text appended to the prior-design prompt}
\textbf{Revision phase (this session replaces the prior-set instructions above)}

This skill already has three policies, \texttt{\{skill\}\_\_h01} to \texttt{\{skill\}\_\_h03}, designed and trained in an earlier session. They are immutable and remain in the deployed library. Each was validated in distribution: it was started from the entry state of its own training segment in each of the 12 training demonstrations (reached by replaying the demonstrated actions exactly), acted for 1.5 times the segment length with its training call arguments at the corresponding demonstration time, and was scored at the final state by the fixed exit rule shown in \texttt{read\_validation}. These are the only evaluation results available to you; no out-of-distribution result exists for you.

Your task in this session:

1. Read the skill (\texttt{read\_skill}), the implementation documents (\texttt{read\_public}: \texttt{INTERFACE.md}, \texttt{TRAINING\_DETAILS.md} and \texttt{panda\_kinematics.py}), the validation results (\texttt{read\_validation}) and, as needed, the original packages (\texttt{read\_original\_policy}) and demonstration steps (\texttt{read\_steps}).

2. Submit a structured report (\texttt{submit\_report}): observed facts supported by the validation results or the demonstrations, observations per original policy, unconfirmed inferences kept separate from facts, the rationale for the new policy, its expected failure signature, and limitations.

3. Design exactly ONE new policy, \texttt{\{skill\}\_\_h04}: submit a plan with exactly one prior (\texttt{submit\_prior\_plan}), then implement, check (\texttt{check\_policy}) and submit (\texttt{submit\_policy}) its package with \texttt{policy\_id} \texttt{\{skill\}\_\_h04}, following every implementation rule above. The developer trains it from scratch with the same data, seed, recipe and update budget as the originals. It may reuse ideas from the originals and it may turn out worse. It will be validated with the same protocol and added to the library next to the originals.

4. \texttt{finish\_skill}.

The instruction above to propose a set of several complementary priors does not apply in this session. Everything else still applies: interfaces, the Cartesian action interface, the read-only verified \texttt{panda\_kinematics.py} and \texttt{panda\_posture.py} (the latter holds the demonstrated joint configurations the IK uses to resolve the arm's redundant degree of freedom), and the handoff documentation. No edits to the originals, no additional policies, no extra training.
\end{promptbox}

\begin{promptbox}{Exp.\ 2 verification, usage notes}
You deliver the validated library of one manipulation skill to a runtime high-level agent that chooses which policy to invoke. Several policies implement the same skill with different priors; each was validated in distribution from the entry state of its own training segment in the 12 training demonstrations (\texttt{read\_validation} shows the protocol, the per-run end-state measurements and the fixed exit rule). No out-of-distribution result exists for you. Read the validation and, as needed, each policy's documents (\texttt{read\_policy}). Then \texttt{submit\_final\_report} with, for every policy, concise usage guidance for the runtime agent (when it is reliable, when to avoid it, how to recognize its failure, how it hands off), grounded only in the validation measurements and the documents, and a short summary. Do not claim results that are not in the validation data.
\end{promptbox}

\subsection{Exp.\ 2 runtime agent}
\label{app:prompt_exp2_runtime}
Full \methodname{} and the ablations with a runtime agent use this prompt unchanged; the ablations differ only in the description fields their catalogue exposes (Appendix~\ref{app:exp2_training}).

\begin{promptbox}{Exp.\ 2 runtime agent}
Complete the supplied geometric goals using the frozen learned policy library. The catalogue exposes each policy's published call schema and semantic purpose. Before first use, read its actual input/output contract, \texttt{PRIOR.md}, \texttt{HANDOFF.json} and measured support. Select the policy and supply the semantic arguments it accepts.

\textbf{Bind your intended operation to a callable policy}

Decide which object or other entity the next operation concerns, the desired goal and any other declared parameters. Express that intent through the policy's typed arguments, not only through reason or notebook text. Different policies may accept different schemas; do not assume a common target parameter or send an undeclared field. Use valid scene references where supported. Numerical goals are allowed only where the contract defines their units, frame, meaning and range.

The executor supplies fresh observations and real causal history to the published input builder. A latched object reference identifies the same object throughout a call while its pose is updated at each replan. A numeric fixed-world target remains fixed if that is its documented meaning. You do not supply substitute measured state, fabricated contact, neural tensors or low-level actuator commands.

Distinguish requested target, possible held object and destination. A policy requested to acquire the next object may first need to release its predecessor, if its documented training support includes that transition. Choose by actual behavior and supported inputs rather than by a promising policy name alone.

\textbf{Observe progress and control transitions}

Use current object motion, height, finger configuration, TCP-object relation and task geometry together. A closed gripper, a reached pose or a predicted phase alone does not establish contact or completion. If observations admit several explanations, choose a useful supported action and a stopping condition that makes the relevant transition observable. Perfect certainty or an unavailable contact sensor is not required to act.

Every published policy call accepts execution duration, numeric \texttt{stop\_when} groups, reason and notebook alongside its own arguments. All conditions in a group must hold; any matching group returns control after a physical step. Use only exposed metrics and comparisons. absolute means the current metric; \texttt{invocation\_start} means current minus that call's initial metric. Check for conditions already true at entry. A stopped call is an opportunity to reassess, not proof of a successful skill. An empty stop list runs the requested duration unless the task succeeds.

Choose durations that permit useful motion and expose uncertain contact or handoff transitions before running past them. Inspect returned state, goal predicates and \texttt{metric\_ranges}. An object low at the end might have lifted earlier; extrema do not imply simultaneous conditions. Use actual states and further supported observations for the decision that matters.

The executor maintains causal observation history across policy calls. A policy or argument change discards incompatible pending actions and replans with the new binding. Consecutive calls with the same policy and arguments retain the valid action stream according to its published execution contract. A shorter invocation is not automatically a fresh sample. Changing a target requires an actual parameter change and appropriate training support.

\textbf{Continue or recover with evidence}

After a failed attempt, state the observed change, the failed expectation and why continuing or selecting another policy/argument could help. Continue a useful transition; choose another supported behavior when the current one no longer fits. Avoid repeatedly cycling through choices without updating the explanation from outcomes. One failed call does not establish that the library is exhausted.

Recovery outside demonstrated support is a hypothesis that can be attempted and evaluated. Revise the plan when observations contradict it. Continue while a credible route remains; finish honestly when you judge the available capabilities cannot complete the task. Do not cite an unknown remaining execution budget.

Preserve achieved goals where feasible while reasoning about handoff readiness. Final success is exactly the supplied whole-task predicate, with no extra release, clearance, speed or hold requirement.

Maintain a concise notebook in your own words: achieved goals, current physical state and bindings, important attempts and outcomes, unresolved facts and the next intended progress. Use the exposed observation/history tools when details are missing. Observation and API deliberation do not advance physics. You cannot modify weights, schemas or adapter code during evaluation or invoke capabilities absent from the published catalogue.
\end{promptbox}

\subsection{SinglePrior baseline}
\label{app:prompt_singleprior}
The SinglePrior designer receives the prior-design prompt adapted to one full-task policy. It opens with the paragraph below, has no segmentation, handoff overlap, or runtime arguments, and requires its single prior to target displacement of the objects the task manipulates or approaches. The full text will be released with the code.

\begin{promptbox}{SinglePrior baseline (opening paragraph)}
You are the offline scientific designer of ONE full-task Single Prior Diffusion Policy. Read the complete original training demonstrations and the task contract, independently choose your inductive bias, then implement, check and submit one immutable model. There is no segmentation, runtime HL, policy selection, stage label or caller-selected object. Use only causal observed state to decide what to do. \texttt{call\_args\_schema} must be \{"type":"object","properties":\{\},"required":[],"additionalProperties":false\}. Each original demonstration must have exactly one complete training binding with \texttt{call\_args}=\{\}. No previous agent hypothesis is supplied. The assignment statement describes scope only. For \texttt{HANDOFF.json} document full-task entry/exit and limitations; it is not a runtime controller.
\end{promptbox}

\clearpage
\section{Extended Related Work}
\label{app:related}
This appendix complements Section~\ref{sec:related_work}.
Appendix~\ref{app:related_roles} summarizes how related families use task structure, Appendix~\ref{app:related_more} discusses further work in each family, and Appendix~\ref{app:related_compare} compares \methodname{} with the closest methods.

\subsection{Two roles of task structure}
\label{app:related_roles}
Table~\ref{tab:related_roles} contrasts structure that shapes what a low-level policy learns with structure that tells a high-level decision maker what the policy does.
Most families address one role; shared symbolic abstractions address both, but through a fixed rule or specification.

\begin{table}[h]
\centering\small
\caption{Two roles of task structure across related families.}
\label{tab:related_roles}
\begin{tabularx}{\linewidth}{@{}>{\raggedright\arraybackslash}p{0.2\linewidth}YYY@{}}
\toprule
Family & Shapes the policy & Informs the caller & Structure designed by \\
\midrule
Structural priors & Yes & No; the prior remains in training & A human, per task family \\
Agents designing learning & Yes, one design per policy & No; the design serves training only & An agent \\
Generalist VLAs & Implicitly, through data & An instruction without stated scope & Data \\
Agents calling tools & No; tools are given & Names, feasibility estimates, or execution records & Designers or pretraining \\
Shared abstractions & Partly, through a fixed rule or recipe & Predicates, preconditions and effects, or skill specifications & A designer or a fixed rule \\
\methodname{} & Yes, with alternative priors per skill & The same priors, with handoff overlap and training support & An agent, per skill \\
\bottomrule
\end{tabularx}
\end{table}

\subsection{Additional related work}
\label{app:related_more}
\paragraph{Structural priors.}
Some priors are specific to one operation, such as articulation flow for doors and drawers and screw motions for bimanual tasks~\citep{eisner2022flowbot3d,bahety2024screwmimic}. Evidence that the right structure depends on the task includes studies of wrist-camera views and 3D point clouds, whose benefits hold only under specific conditions~\citep{hsu2022vision,ling2023efficacy}. In addition, an incorrect symmetry assumption bounds achievable accuracy~\citep{wang2023general}, and the gain from a single visual-cue prior differs across task suites~\citep{dai2025aimbot}.
CodeDiffuser instead generates task-specific attention maps with VLM-written code to condition a diffusion policy~\citep{yin2025codediffuser}.

\paragraph{Agents and robot tools.}
Skill-level interfaces include geometric feasibility checks over skill sequences and predicate invention for learned skills~\citep{lin2023text2motion,yang2025skillwrapper}, and handoff-aware memory records entry conditions discovered by exploration~\citep{xu2026baton}.

\paragraph{Symbolic abstractions and skill discovery.}
Temporal abstraction and TAMP foundations include options, symbols derived from skills, and sampler-based planning~\citep{sutton1999between,konidaris2018skills,garrett2020pddlstream}.
Abstractions for planning are learned as operators, compositional skill models, and predicates, including predicates proposed from pretrained VLMs~\citep{silver2021learning,silver2023neurosymbolic,silver2023predicate,liang2025visualpredicator}; competence-aware planning learns when skills succeed~\citep{kumar2024practice}.
Skills with symbolic interfaces are learned through planner-guided reinforcement learning~\cite{cheng2023league} or from demonstrations~\citep{liu2025blade,quartey2026pacts}, and skills and structure are discovered from demonstrations~\citep{zhu2022buds,wan2024lotus}.
Planners and language models sequence learned or human-guided skills~\citep{mandlekar2023hitltamp,dalal2024psl,agia2023stap}.
Transition policies and skill-chaining methods address handoffs between skills~\citep{konidaris2009skill,lee2019composing,lee2022tstar,mishra2023gsc}; \methodname{}'s overlapping training segments similarly widen a successor skill's initiation set. MaestroMotif and SCALAR use language skill specifications for both training and composition, but these specifications describe skill effects rather than inductive bias~\citep{klissarov2025maestromotif,zabounidis2026scalar}.

\paragraph{Agents that design learning systems.}
Further systems shape rewards, generate simulation tasks, and design sim-to-real rewards and domain randomization~\citep{xie2024text2reward,wang2024robogen,wang2024gensim,ma2024dreureka}.
Language models also compose encoder architectures, write policy structures from domain knowledge, and automate reinforcement-learning pipelines~\citep{wang2024lesr,yu2025lacer,zhu2025kim,wei2025agent2}; they also run policy-improvement loops~\citep{xiao2026enpire,khandelwal2026agent,yu2026agentic}.
When learned skills are reused by a later decision maker, they are described by names, code, or effects~\citep{wang2024voyager,klissarov2025maestromotif,zabounidis2026scalar,nie2026teach}; Agentic Skill Discovery retains several policy variants per skill and leaves choosing among them to future work~\citep{zhao2026agentic}.

\subsection{Comparison with the closest methods}
\label{app:related_compare}
Table~\ref{tab:related_compare} compares \methodname{} with methods that come closest to either role.
The comparison concerns method design; the settings differ, ranging from real robots to simulated manipulation, NetHack, and Craftax, and the table implies no empirical ranking.

\begin{table}[h]
\centering\small
\setlength{\tabcolsep}{3pt}
\caption{Design comparison with the closest methods. ``Reads'' is what the high-level decision maker sees about each low-level policy before or during invocation.}
\label{tab:related_compare}
\begin{tabularx}{\linewidth}{@{}>{\raggedright\arraybackslash}p{0.2\linewidth}YYYYY@{}}
\toprule
Method & Low-level policies & Structure designed by & Alternatives per skill & Reads & Decision maker \\
\midrule
LGA~\citep{peng2024lga} & Imitation from few demonstrations & A language model, one state abstraction per task & No & Nothing; the abstraction is the policy input & None \\
KALM~\citep{fang2025kalm} & Keypoint-conditioned imitation & A VLM proposes keypoints checked on demonstrations & No & Nothing & None \\
SymSkill~\citep{shao2025symskill} & Dynamical-system skills in relative frames & Co-invented predicates; a VLM selects frames offline & No & Predicates and operators & Symbolic planner \\
\citet{lorang2025fewshot} & Diffusion controllers & A fixed rule restricts inputs to operator-relevant objects & No & Learned operators & Symbolic planner \\
DR-LfD~\citep{chen2026drlfd} & Diffusion policies and equivariant primitives & A fixed rule based on contact complexity & No & Learned initiation and termination keyposes & Task and motion planner \\
MaestroMotif~\citep{klissarov2025maestromotif} & Reinforcement-learned skills in NetHack & Human skill descriptions; LLM-derived rewards & No & Skill descriptions, initiation and termination code & LLM-written policy over skills \\
Agentic Skill Discovery~\citep{zhao2026agentic} & Reinforcement-learned skills in simulation & An LLM proposes tasks and rewards & Yes; selection left open & Skill names or descriptions & LLM \\
RoboClaw~\citep{li2026roboclaw} & Fine-tuned $\pi_{0.5}$ & A fixed recipe; the agent chooses data & No & Subtask instructions and tool outputs & VLM agent \\
Harness VLA~\citep{zhang2026harness} & Frozen VLA and analytic primitives & None; policies are pretrained & No & Memory of successful calls & Coding agent \\
BATON~\citep{xu2026baton} & Frozen VLA and analytic primitives & None; policies are pretrained & No & Readiness and handoff conditions found by exploration & Coding agent \\
RoboHarness~\citep{huang2026roboharness} & Pre-existing heterogeneous policies & None; policies are given & No & Policy cards updated from execution & Coding agent \\
\midrule
\methodname{} & Diffusion policies from 12 demonstrations per task & The construction agent, per skill & Yes; three per skill in Exp.~2, plus one after verification & Prior, handoff overlap, measured training support, and verification results & Runtime agent \\
\bottomrule
\end{tabularx}
\end{table}

\paragraph{Agent-designed priors.}
LGA and KALM show that language and vision-language models can choose representation priors for imitation learning, and ATK selects task-driven keypoints automatically~\citep{peng2024lga,fang2025kalm,zhang2025atk}.
Each automates one family of priors and produces one design per task, which the policy consumes as input.
\methodname{} does not claim agent-designed priors by themselves as new.
Its construction agent proposes priors from several families for each skill, keeps the resulting policies as alternatives, and reuses each prior as runtime documentation.

\paragraph{SymSkill.}
SymSkill is the closest precedent to \methodname{}~\citep{shao2025symskill}.
Offline, it jointly learns predicates, operators, and goal-oriented skills from unlabeled, unsegmented demonstrations, using relative frames, with a VLM assisting the choice of reference frames, and stable dynamical-system policies.
Online, a symbolic planner composes and reorders skills to reach goals stated as conjunctions of learned predicates, and the system recovers from failures at both the motion and symbolic levels in real time.
Both methods learn reusable skills from complete demonstrations and let structure found during learning organize long-horizon composition.
They differ in three ways.
First, SymSkill's structure follows one design family, whereas \methodname{}'s construction agent proposes priors from different families for each skill and implements them as representations, architectures, or losses.
Second, SymSkill associates each operator with a learned skill policy, whereas \methodname{} keeps several policies with different priors for the same skill and lets the runtime agent choose among them.
Third, SymSkill's planner reads symbolic preconditions and effects, whereas \methodname{}'s runtime agent reads natural-language priors, handoff overlap, and measured training support, and sets a duration and stop conditions for each call.
The approaches are complementary, since SymSkill's learned predicates and reactive recovery could provide execution checks and fallback behaviors within \methodname{}'s interfaces.

\paragraph{Shared abstractions.}
Beyond SymSkill, \citet{lorang2025fewshot} and DR-LfD~\citep{chen2026drlfd} also let one structure serve both learning and planning.
In each, a fixed rule or template determines how the abstraction constrains a controller, and each skill receives one learning recipe.
MaestroMotif and SCALAR share language specifications between training and composition~\citep{klissarov2025maestromotif,zabounidis2026scalar}, but these specifications describe what a skill achieves and when it starts or stops, rather than the inductive bias or training support of its policy.

\paragraph{Agentic harnesses.}
Harness VLA, BATON, and RoboHarness invoke policies with an instruction, a step budget, and a stop condition, as \methodname{} does; \methodname{} does not claim this call as new.
These systems learn operating ranges, entry conditions, or policy cards from executions after the policies exist.
\methodname{} derives its documentation from the construction-time prior and from the overlap built into training data, so this information is available before the first execution.
RoboClaw keeps one agent across data collection, training, and deployment, but its agent selects data for a fixed training recipe rather than structural priors.

\paragraph{Alternatives and handoffs.}
Agentic Skill Discovery retains several policy variants per skill and leaves the choice among them to future work~\citep{zhao2026agentic}; in \methodname{}, the runtime agent makes this choice using the priors that distinguish the variants.
\methodname{}'s overlapping training segments follow work that widens a successor skill's initiation set~\citep{konidaris2009skill,lee2022tstar}, and we treat the overlap as a design choice rather than a contribution.

\end{document}